\documentclass[11pt,reqno]{amsproc}
\usepackage[T1]{fontenc}
\usepackage{amsmath,amscd,amssymb}
\usepackage{cite}
\usepackage{color}
\usepackage{graphicx}
\usepackage{psfrag,epsfig}
\usepackage{multirow}
\usepackage{rotating}
\usepackage[letterpaper,margin=0.5in]{geometry}
\usepackage{subfigure}
\usepackage{enumerate}
\usepackage{comment}
\usepackage{lineno}
\usepackage{algorithmic}
\usepackage{algorithm}
\usepackage[scriptsize]{caption}
\usepackage{bbold}
\usepackage{tablefootnote}
\usepackage[debug=false, colorlinks=true, pdfstartview=FitV, linkcolor=blue, citecolor=blue, urlcolor=blue]{hyperref}
\usepackage{url}
\numberwithin{equation}{section}
\usepackage{chemarr}
\usepackage[version=3]{mhchem}
\newcommand*\patchAmsMathEnvironmentForLineno[1]{%
  \expandafter\let\csname old#1\expandafter\endcsname\csname #1\endcsname
  \expandafter\let\csname oldend#1\expandafter\endcsname\csname end#1\endcsname
  \renewenvironment{#1}%
     {\linenomath\csname old#1\endcsname}%
     {\csname oldend#1\endcsname\endlinenomath}}%
\newcommand*\patchBothAmsMathEnvironmentsForLineno[1]{%
  \patchAmsMathEnvironmentForLineno{#1}%
  \patchAmsMathEnvironmentForLineno{#1*}}%
\AtBeginDocument{%
\patchBothAmsMathEnvironmentsForLineno{equation}%
\patchBothAmsMathEnvironmentsForLineno{align}%
\patchBothAmsMathEnvironmentsForLineno{flalign}%
\patchBothAmsMathEnvironmentsForLineno{alignat}%
\patchBothAmsMathEnvironmentsForLineno{gather}%
\patchBothAmsMathEnvironmentsForLineno{multline}%
}
\usepackage{siunitx}
\DeclareSIUnit{\mEq}{mEq}
\title[Deep learning for watershed modeling]{SWAT Watershed Model Calibration using Deep Learning}
\author[M.~K.~Mudunuru et.al.,]{M.~K.~Mudunuru$^{*}$, K.~Son, P.~Jiang, X.~Chen \\
{\small Watershed \& Ecosystem Science, Pacific Northwest National Laboratory, Richland, WA 99352, USA}
}
\thanks{$^*$Corresponding author's email address:~\texttt{maruti@pnnl.gov}}
\date{\today}
\begin{document}
\maketitle
%
%
\section*{ABSTRACT}
Watershed models such as the Soil and Water Assessment Tool (\texttt{SWAT}) consist of high-dimensional physical and empirical parameters.
These parameters need to be accurately calibrated for models to produce reliable predictions for streamflow, evapotranspiration, snow water equivalent, and nutrient loading.
Existing parameter estimation methods are time-consuming, inefficient, and computationally intensive, with reduced accuracy when estimating high-dimensional parameters.
In this paper, we present a fast, accurate, and reliable methodology to calibrate the \texttt{SWAT} model (i.e., 21 parameters) using deep learning (DL). 
We develop DL-enabled inverse models based on convolutional neural networks to ingest streamflow data and estimate the \texttt{SWAT} model parameters.
Hyperparameter tuning is performed to identify the optimal neural network architecture and the nine next best candidates.
We use ensemble \texttt{SWAT} simulations to train, validate, and test the above DL models.
We estimated the actual parameters of the \texttt{SWAT} model using observational data.
We test and validate the proposed DL methodology on the American River Watershed, located in the Pacific Northwest-based Yakima River basin.
Our results show that the  DL models-based calibration is better than traditional parameter estimation methods, such as generalized likelihood uncertainty estimation (GLUE).
The behavioral parameter sets estimated by DL have narrower ranges than GLUE and produce values within the sampling range even under high relative observational errors.
This narrow range of parameters shows the reliability of the proposed workflow to estimate sensitive parameters accurately even under noise.
The best DL-based calibrated set has $R^2$, Nash-Sutcliffe efficiency, logarithmic Nash-Sutcliffe efficiency, and Kling-Gupta efficiency scores of 0.53, 0.67, 0.78, and 0.74, respectively.
The best GLUE-based calibrated set has 0.48, 0.6, 0.7, and 0.68, respectively.
The above scores show that the DL calibration provides more accurate low and high streamflow predictions than GLUE.
Due to its fast and reasonably accurate estimations of process parameters, the proposed DL workflow is attractive for calibrating integrated hydrologic models for large spatial-scale applications.
\newline
\newline
\textbf{KEYWORDS:}~\texttt{SWAT},
calibration,
watershed modeling,
parameter estimation,
inverse problems,
convolutional neural networks.

\section*{\textbf{HIGHLIGHTS AND NOVELTY}}
\begin{itemize}
    \item We developed a deep learning (DL) methodology to estimate the parameters of the \texttt{SWAT} model.
    \item  DL-enabled \texttt{SWAT} model calibration shows higher streamflow prediction accuracy than traditional parameter estimation methods such as Generalized Likelihood Uncertainty Estimation (GLUE).
    \item Estimated \texttt{SWAT} model parameters from observational discharge are within the sampling range of ensemble simulations and are well clustered even under noise.
    \item An added benefit is that DL-enabled parameter estimation after training is at least $\mathcal{O}(10^3)$ times faster than GLUE-based methods, which may require 100-1000s of forward model runs.
\end{itemize}

\section{INTRODUCTION}
\label{Sec:S1_Intro}
Watershed models are frequently used to estimate streamflow and water exchanges among different components of the terrestrial water cycle.
These components are affected by a wide range of anthropogenic activities (e.g., agricultural intensification) and climate perturbations (e.g., wildfire, rain-on-snow, rising temperatures and precipitation, earlier snowmelt in mountainous regions). \cite{singh2010watershed,singh2003watershed,daniel2011watershed}.
Watershed models can also assess the sustainability of the water supply for effective water resource management.
Some popular and open-source watershed modeling software that can accurately simulate various components of water cycling in intensively managed watersheds include the Soil and Water Assessment Tool \texttt{SWAT} and its variants (e.g., \texttt{SWAT-MRMT-R}) \cite{neitsch2011soil,douglas2010soil,SWAT-MRMT-Fang}, the Advanced Terrestrial Simulator (\texttt{ATS}) \cite{ATS-2020}, the Precipitation Runoff Modeling System (\texttt{PRMS}) \cite{leavesley1983precipitation,markstrom2015prms}, the Weather Research and Forecasting Model Hydrological modeling system (\texttt{WRF-Hydro}) \cite{sampson2018wrf,WU2021104959}, \texttt{RHESSys} \cite{tague2004rhessys}, \texttt{VIC} \cite{hamman2018variable}, \texttt{MIKE-SHE} \cite{graham2005flexible}, \texttt{DHSVM} \cite{cuo2008hydrologic}, and \texttt{HSPF} \cite{donigian1995hydrological}.

The software mentioned above has process models that simulate the watersheds' different hydrological components (e.g., infiltration, groundwater flow, streamflow). 
Watershed hydrologic models feature two types of parameters \cite{johnston1976parameter,mein1978sensitivity}.
The first type includes parameters with physical significance (e.g., permeability, porosity) \cite{nakshatrala2019interface} that can be determined from observational data.
The second type of parameters is conceptual or empirical, which are currently impossible or difficult to measure directly.
Most of the watershed simulators mentioned above (e.g., \texttt{SWAT}, \texttt{PRMS}) consist of parameters that fall in the second category \cite{singh2010watershed}.
As a result, observational data, such as streamflow collected at the watershed outlet, are used to estimate the conceptual parameters through model calibration.
Many watershed models, including the \texttt{SWAT} and \texttt{PRMS} models, can only achieve adequately accurate predictions after calibrating their parameters to available observations, making them less ideal for ungauged watersheds.
On the other hand, advanced mechanistic process-based watershed models (e.g., \texttt{ATS}) can predict watershed responses with reasonable accuracy without undergoing intensive model calibration \cite{cromwell2021estimating}.
The parameters in these mechanistic models are measurable and physically significant.

Various techniques and software for calibrating watershed models exist in the literature \cite{duan2004calibration}. 
Popular methods include generalized likelihood uncertainty estimation (GLUE) \cite{blasone2008generalized,nott2012generalized}, maximum likelihood estimation \cite{myung2003tutorial}, the shuffled complex evolution method developed at the university of Arizona (SCE-UA) \cite{duan1994optimal}, Bayesian parameter estimation methods \cite{thiemann2001bayesian,misirli2003bayesian,gupta2003advances}, ensemble-based data assimilation methods (e.g., ensemble Kalman filter, ensemble smoother) \cite{evensen1994sequential,van1996data,evensen2003ensemble,chen2013application,evensen2018analysis,jiang2021dart}, and adjoint-based methods \cite{aster2018parameter,tarantola2005inverse}. 
These techniques underpin popular software packages such as \texttt{PEST} \cite{doherty2010approaches}, \texttt{DAKOTA} \cite{adams2009dakota}, \texttt{SWAT-CUP} \cite{abbaspour2013swat}, \texttt{MATK} \cite{MATK-v1}, \texttt{MADS} \cite{MADS-v1}, and \texttt{DART} \cite{anderson2009data}, which are developed to perform model calibration. 
These existing calibration methods and tools are time-consuming (e.g., can have slow convergence), require good initial guesses, and can be computationally intensive (e.g., may require a lot of forward model runs or running on high-performance computing clusters) \cite{zhang2016moving,bacu2017swat,rouholahnejad2012parallelization}. 
Moreover, calibration using such tools can potentially result in reduced accuracy when estimating high-dimensional parameters ($>10$) due to the difficulty in solving a multi-objective optimization problem \cite{duan2004calibration,eckhardt2005automatic}.
Many of the methods mentioned above have challenges (see supplementary text S1) estimating the strong nonlinear relationships between parameters and observational data \cite{franco2017multi}. 
Recent advances in deep learning (e.g., deep neural networks, convolutional neural networks) show promise for developing reliable model calibration workflows that overcome the above-mentioned challenges \cite{cromwell2021estimating,gabrielli2017introducing}.
In this paper, we present a deep learning (DL) methodology for estimating high-dimensional \texttt{SWAT} model parameters from observational streamflow data efficiently, reliably, and with reasonably good accuracy.

Deep learning (DL) shows promise in aiding conceptual or process-based inverse modeling associated with highly nonlinear relationships \cite{gabrielli2017introducing,afzaal2020groundwater,marccais2017prospective,nearing2021role,zhang2009approximating,sit2020comprehensive}.
DL utilizes multiple neural layers to extract features that are representative of inputs.
DL-enabled inverse models for parameter estimation are known to be robust with observational errors or noise \cite{edwards2018deep,rolnick2017deep,gupta2019dealing,rudi2020parameter}.
In hydrology, neural networks (e.g., deep, convolutional, recurrent) have been used to model and predict rainfall-runoff, streamflow, water quality, and precipitation \cite{shen2018transdisciplinary,khandelwal2020physics,bhasme2021enhancing} based on synthetic data.
Here synthetic data refers to data generated through watershed model simulations conditioned on site-specific parameters.
Recently, deep neural networks guided by physics and domain knowledge have been used to predict lake temperatures, streamflows, and estimate biophysical parameters \cite{willard2020predicting,jia2021physics,read2019process,rahmani2021exploring,dagon2020machine}.
This study uses convolutional neural networks (CNNs), which are popular in hydrological applications \cite{sadeghi2019persiann,van2020deep,jagtap2021deep}. 

CNNs offer many advantages over dense neural networks (DNNs).
A significant advantage of CNNs is that they learn local representations (or patterns) in time-series data while the DNNs can only understand global features.
CNNs generally have lower data requirements than DNNs to obtain similar performance levels due to fewer trainable weights.
This superior performance of CNNs can be attributed to the multiple convolutional layers learning hierarchical patterns from the inputs. 
The resulting broader set of abstract patterns are used to develop nonlinear mappings between streamflow and \texttt{SWAT} model parameters.
Another benefit of DL-enabled inverse models is their low inference time for parameter estimation compared to traditional methods.
However, the data requirements and associated training time needed to develop such inverse models can be substantial.
Once the DL-enabled inverse model is trained on synthetic data, it can allow for seamless assimilation of observational data, thereby significantly reducing the time required to estimate parameters in high-dimensional space \cite{cromwell2021estimating}.

\subsection{Main contributions}
\label{SubSec:Main_Contribution}
The main contribution of this study is developing a fast and accurate parameter estimation methodology using deep learning that calibrates watershed models better than traditional methods.
The DL-enabled inverse models are built on synthetic data generated using the \texttt{SWAT} model and guided by sensitivity analysis.
Hyperparameter tuning is performed to identify the top ten architectures based on validation mean squared error.
In many scenarios, observational data is most likely corrupted by noise.
We analyze the influence of noise on parameter estimation in both testing and observation data.
A significant advantage of the proposed DL workflow is that it accurately estimates sensitive parameters even at high noise levels (e.g., 25\% relative observational errors).
Moreover, these estimated parameters are well clustered and within the prior sampling range, showing the proposed methodology's robustness to observational errors.
Compared to the GLUE method, the parameters estimated by the DL-enabled inverse models provide more accurate streamflow predictions within and beyond the calibration period.
The GLUE method identified a set of behavioral parameters within the ensemble parameter combinations.
By `behavioral' parameters, we mean to signify parameter sets whose \texttt{SWAT} model simulations are deemed to be `acceptable' upon satisfying certain user-defined performance metrics (e.g., Kling-Gupta efficiency greater than 0.5) on observational data \cite{blasone2008generalized}.
Based on a cutoff threshold that utilizes metrics such as Kling-Gupta efficiency, the entire set of simulations is then split into behavioral and non-behavioral parameter combinations.
The behavioral parameter set provides better accurate predictions than the non-behavioral set.
Our analysis also showed that the DL estimated set clusters better than the GLUE-based behavioral set.
Another advantage of the proposed DL-based inverse models is that it is at least $\mathcal{O}(10^3)$ times faster than the GLUE-based method. 
From a computational cost perspective, traditional parameter estimation (e.g., using \texttt{PEST}, \texttt{DAKOTA}) requires multiple forward model runs.
As a result, inversion modeling requires high-performance computational resources that can be prohibitively expensive.
The savings in computational cost enables our DL-enabled parameter estimation to be inclusive (i.e., easy to adapt using transfer learning \cite{zhuang2020comprehensive}) and ideal for calibrating multi-fidelity models (e.g., \texttt{ATS}, \texttt{PFLOTRAN}, \texttt{WRF-Hydro}, \texttt{PRMS}) at spatial scales of watersheds and basins.

\subsection{Outline of the paper}
\label{SubSec:Outline_Paper}
The paper is organized as follows: Sec.~\ref{Sec:S1_Intro} discusses state-of-the-art methods for parameter estimation and their limitations.
We also demonstrate the need for developing DL workflows to better calibrate hydrological models, such as the \texttt{SWAT}. 
Section~\ref{Sec:S2_SWAT_Model} describes the study site and \texttt{SWAT} model developed using a National Hydrography Dataset PLUS (NHDPLUS v2)-based watershed delineation.
We discuss data generation to develop DL-enabled inverse models and provide the top seven parameters that are sensitive to discharge. 
We also compare observation data with the \texttt{SWAT} model ensemble simulations. 
Section~\ref{Sec:S3_DNN_Calib} introduces the proposed DL methodology for estimating \texttt{SWAT} parameters.
We developed three different types of DL-enabled inverse models using sensitivity analysis as a guide.
We performed hyperparameter tuning to identify the optimal deep learning model architectures and described the associated computational costs for training the DL models and generating inferences (e.g., on test and observational data).
Sec.~\ref{Sec:S4_Results} presents the training, validation, and testing results of the DL models.
We compare the performance of DL estimated parameters with that of the GLUE method.
Performance of calibration model within and beyond calibration period is provided.
Finally, Sec.~\ref{Sec:S5_Conclusions} presents our conclusions.

\section{STUDY SITE and DATA GENERATION}
\label{Sec:S2_SWAT_Model}
This section first describes the study site, the American River Watershed (ARW) in the Yakima River Basin (YRB), before discussing the \texttt{SWAT} model, its parameters, and specifics on the ensemble runs needed to develop DL-enabled inverse models.
We also compare the observational streamflow/discharge used to calibrate the \texttt{SWAT} model with the ensemble runs within the calibration period (i.e., from 10/01/1999 to 30/09/2009).

\subsection{Study site}
\label{SubSec:S2_Study_Site}
The Yakima River Basin (YRB), situated in Eastern Washington, has a drainage area of about 16,057 $\mathrm{km}^2$ \cite{mastin2002watershed}.
The YRB produces an average annual runoff of approximately 90 $\mathrm{m}^3 \mathrm{s}^{-1}$. 
A major tributary of the Yakima River is the American River near Nile in Washington, a third-order order stream, with a watershed of about $205 \, \mathrm{km}^2$. 
Within the ARW, the mean annual precipitation and temperature range from 978 to 2164 mm and 2.8 to 4.9$^{\mathrm{o}}\mathrm{C}$, respectively.
This range is estimated according to the 30 years normalized PRISM data \cite{daly2000high,daly2013prism,PRISMdata}.
ARW's climate has a strong seasonality, including cold, wet winters and hot, dry summers. 
About 60\% of precipitation occurs in the winter as snow, with snowmelt occurring from April to June the following year.
Peak snow accumulation and flow occur in April and May, respectively. 
This prior site-specific knowledge shows that the snow process parameters in the \texttt{SWAT} model are essential.
Guided by the information mentioned above and sensitivity analysis, our results demonstrate that we can better estimate such important process model parameters using DL than GLUE.

The ARW has a steep slope that varies from $0^{\mathrm{o}}$ to $83^{\mathrm{o}}$, with a mean slope of $23^{\mathrm{o}}$. 
The major surface geology types are andesitev (72\%), granodiorite(20\%), and alluvium (8\%). 
The primary soil texture is gravelly loamy sand with a maximum soil depth of 1524 mm based on USDA State Soil Geographic Data (STATSGO) \cite{schwarz1995state}. 
This soil is classified as hydrologic group B with moderate runoff potential and infiltration rates. 
Evergreen trees (83\%) and shrub (11\%) dominate land cover and use, with other types of land cover, including urban, grass, and wetlands.
The ARW has a USGS gauging station (USGS 12488500) located in the watershed outlet, recording the daily observed streamflow from 07/16/1988 to the present. 
A snow telemetry (SNOTEL) station (site name:~Morse lake) is located northwest of the watershed.
This SNOTEL station has measured the snow water equivalent, daily precipitation, and air maximum/mean/minimum temperatures from 10/1/1979 to the present.   

\subsection{Brief description of the SWAT model}
\label{SubSec:S2_SWAT_description}
\texttt{SWAT} is a semi-distributed ecohydrological model.
It can simulate land surface hydrology, soil or plant biogeochemistry, and instream processes \cite{arnold2012swat}. 
The \texttt{SWAT} model requires spatial Geographic Information System (GIS) data to simulate various quantities of interest (e.g., streamflow).
ARW's watershed properties such as topography, land cover, and soil are parameterized using this GIS data.
USGS's 10m digital elevation model (DEM) is used to compute the topographic parameters (e.g., drainage area, slope, slope length) with the ARW's basin and sub-basin boundary and stream networks defined by the National Hydrography Dataset Plus (NHDPus) catchment/streams. 
Previous studies \cite{chiang2015nhdplus,moore2016road} have demonstrated that NHDPlus-based catchment/streams outperformed the modeled streamflows that did not account for such delineation. 

\begin{figure}
  \centering
    \includegraphics[width = 1.05\textwidth]{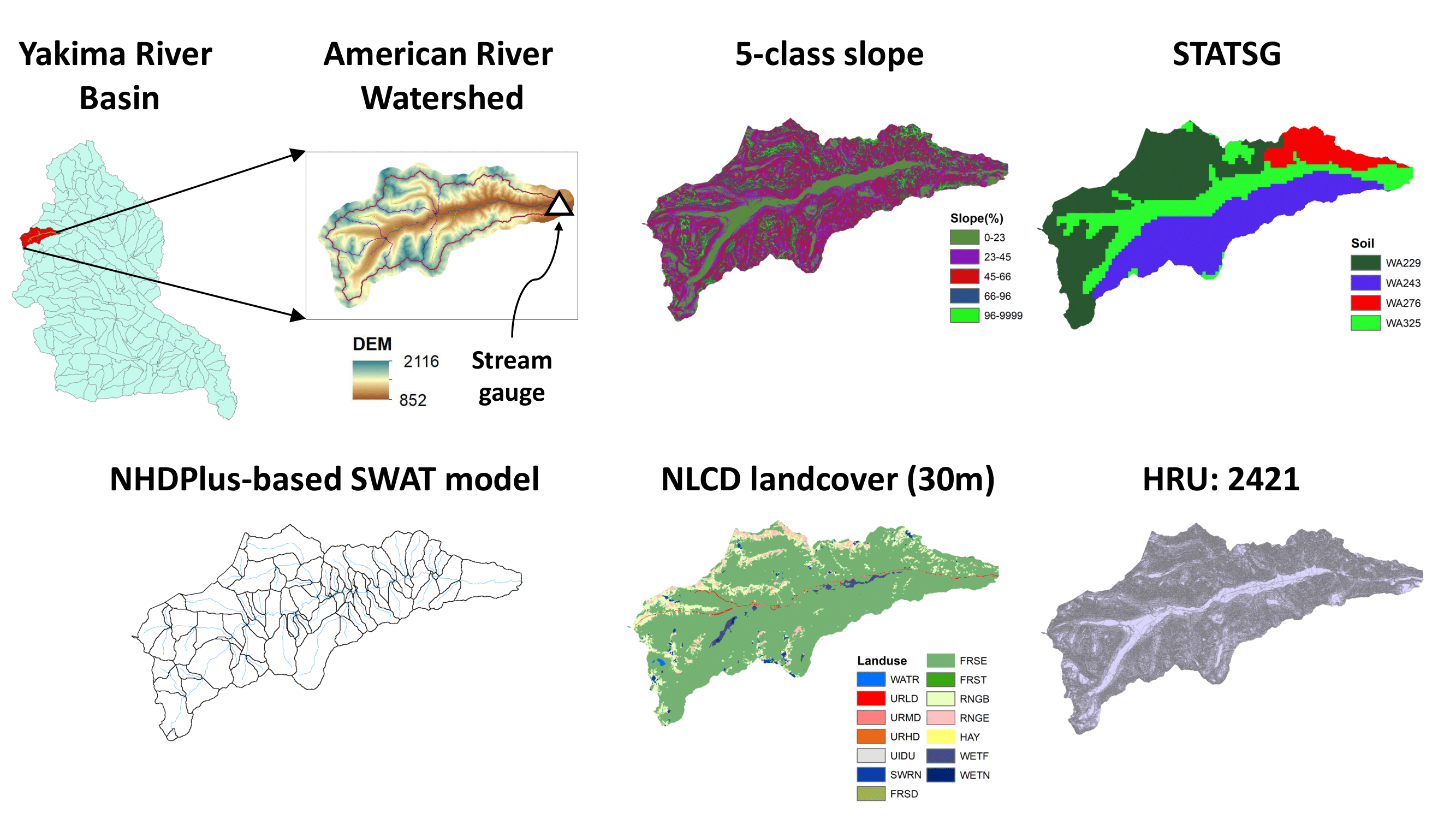}
  \caption{\textbf{NHDPlus-basedSWAT model for ARW:}~This figure shows the watershed delineation and data products used to develop the \texttt{SWAT} model for our study site.
  The top left figures show the delineation of the YRB into different watersheds, including our ARW study site.
  It also shows the DEM used for modeling the ARW and the associated stream gauge at the watershed outlet.
  The top right figures show the spatially varying slope and soil data within the study site.
  The bottom left figure shows the delineation of third-order streams in the ARW. 
  The bottom right figures show the spatially varying land cover (primarily evergreen) and the number of HRU employed in the \texttt{SWAT} model.
  \label{Fig:SWAT_ARW_Study_Site}}
\end{figure}

Figure~\ref{Fig:SWAT_ARW_Study_Site} shows the NHDPlus-based \texttt{SWAT} model used to simulate streamflow at the ARW study site.
This model is composed of 87 sub-basins with five slope classes (percent rise of the slope):~(1) 0-26, (2) 26-51, (3) 51-80, (4) 80-129, and (5) 129-999. 
USGS National Landcover Database (NLCD) 2016 (30m resolution) and USDA STATSGO database are used to estimate the land cover/use and soil parameters, respectively. 
Hydrologic response unit (HRU) maps are developed by combining the slope class, land cover/use, and soil type, resulting in a total of 2421 HRUs for the study site (see Fig.~\ref{Fig:SWAT_ARW_Study_Site}). 
The supplementary material text S2 provides additional details on \texttt{SWAT} model development for our study site.

Daily precipitation, maximum air temperature, and minimum air temperature from a daily Daymet \cite{Daymetdata} with 1 km spatial resolution are used to prepare the climate input data for the \texttt{SWAT} model simulations.
Other climate variables (e.g., radiation, relative humidity, wind speed) are generated using weather generators in the \texttt{SWAT}. 
Five elevation bands (at 100m intervals) per sub-basin are created to better represent the spatial variation of the precipitation and temperature within the region.
The spatially varied lapse rate of precipitation and temperature based on the 1 km Daymet data is then applied to each of the ARW's sub-basins.
Basin averaged lapse rates are used if the Daymet is less than five data points per sub-basin. 
Note that we also verified the Daymet products using SNOTEL's precipitation and temperature data.

\subsection{Data for the SWAT model calibration and sensitivity analysis}
\label{SubSec:S2_Training_Data}
Table~\ref{table:SWAT_Params_stats} summarizes the 21 parameters and their associated sample ranges (e.g., minimum and maximum values) we calibrated in the \texttt{SWAT} model to generate simulation data.
The table clearly shows six groups/types of \texttt{SWAT} model parameters: (1) landscape, (2) soil, (3) groundwater, (4) channel, (5) snow, and (6) plant. 
Each parameter in a specified group is calibrated at different spatial scales.
For example, snow group parameters such as SFTMP and SMTMP represent basin-scale snow processes.
Channel group parameters are at the sub-basin level, and soil/groundwater/plant group parameters represent HRU level spatial variation. 
Even though some parameters (e.g., CANMX) are calibrated at the HRU level, their associated values represent basin-scale average as the initial value of that specified parameter is the same for all HRUs.

\begin{figure}
  \centering
    \includegraphics[width = 0.95\textwidth]{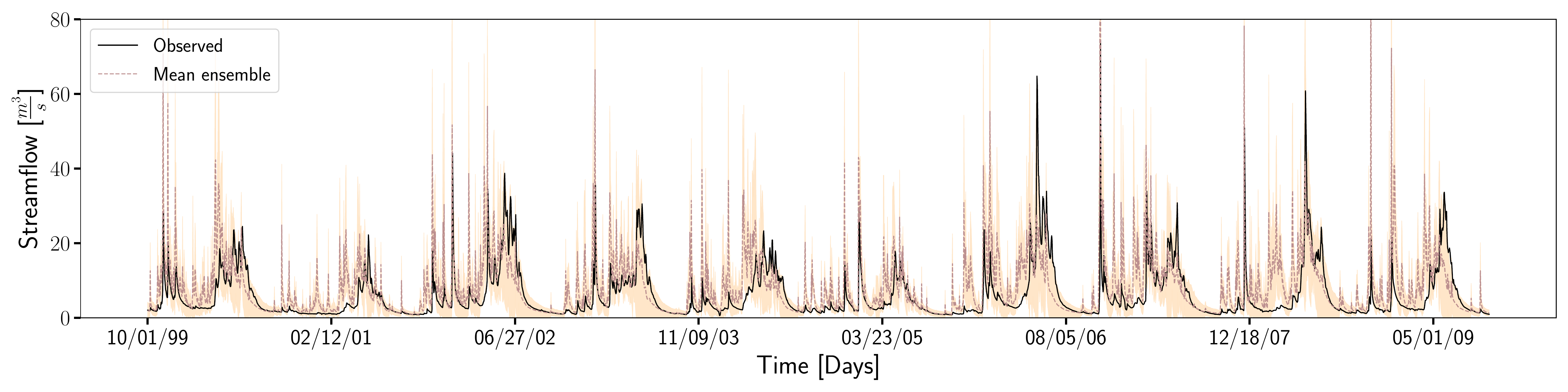}
  \caption{\textbf{SWAT model simulations vs. observational data within calibration period:}~This figure compares the synthetic data generated based on NHDPlus-based \texttt{SWAT} model with observational data for the ARW study site.
  The dark brown color dashed line represents the ensemble mean of 1000 \texttt{SWAT} model simulations.
  The light orange color region represents the standard deviation of ensembles.
  The black-colored line corresponds to the observed streamflow data.
  \label{Fig:SWAT_sim_vs_obsdata}}
\end{figure}

We randomly generated 1000 sets of these 21 parameters using uniform distribution to develop DL-enabled inverse models.
The daily streamflow/discharge data simulated using the \texttt{SWAT} model for these 1000 realizations are shown in Fig.~\ref{Fig:SWAT_sim_vs_obsdata}.
The simulated time for the \texttt{SWAT} model calibration is between 10/01/1999 to 30/09/2009, referred to as the calibration period.
The validation period is beyond 30/09/2009, stretching from 10/01/2009 to 30/09/2016.
The calibrated \texttt{SWAT} model is run during the validation period, and its performance is then compared with the observational data.
Figure~\ref{Fig:SWAT_sim_vs_obsdata} compares the mean ensemble of simulated discharge (i.e., 1000 realizations) with the observational data.
The light orange color represents the standard deviation of the 1000 simulated discharge realizations.
This figure qualitatively shows the similarities of the trends in the simulated discharge and observed data.
Process model assumptions in the \texttt{SWAT} modules contribute to clear structural deficiencies in the predicted streamflow (e.g., over/under predictions of peak/low flows).
Thus, \texttt{SWAT} model fidelity needs enhancement to overcome these structural deficiencies.
The generated data is used to estimate \texttt{SWAT} parameters by both the DL-based calibration and the GLUE method.
The GLUE-based \texttt{SWAT} model calibration is also compared with the observational data for both periods.
The behavioral model parameter sets (i.e., from GLUE) are selected based on KGE metrics.
One can also use the other accuracy measures (e.g., NSE, logNSE, $R^2$-score) to evaluate the calibrated \texttt{SWAT} model performance. 
We note that each of the accuracy metrics reflects a different aspect of the calibration performance.
As a result, including multiple accuracy metrics for evaluating a calibrated model has obvious advantages. 

Table~\ref{table:SWAT_Params_stats} also provides the sensitive parameters that influence simulated discharge at the ARW study site.
Simulated discharge is sensitive to seven of the 21 parameters -- SFTMP, SMTMP, SMFMX, CH\_K2, CH\_N2, ALPHA\_BF, and RCHRG\_DP.
The important parameters mentioned above are identified using mutual information (MI) methods \cite{Cover2006}. Figure S1 in the supplementary material provides the MI-based parameter ranking \cite{herman2017salib,saltelli2008global}. 
We note that discharge is primarily influenced by the SFTMP parameter (i.e., the most sensitive).
This inference from MI is per the seasonality of the study site mentioned in Subsec.~\ref{SubSec:S2_Study_Site}.
As ARW is snow-dominated, sensitivity analysis identified that snow processes influence peak flow behavior.   

\begin{table}[htbp]
    \centering
    \caption{This table provides a list of \texttt{SWAT} model parameters that are calibrated using the proposed DL methodology. 
    The associated lower and upper limits of parameter values are also specified.}
    \scriptsize
    \begin{tabular}{|c|c|c|c|c|c|c|}\hline
        \textbf{Parameter} & \textbf{Parameter\footnotemark} & \textbf{Lower} & \textbf{Upper} & \textbf{Brief description (units)} & \textbf{Parameter} & \textbf{Spatial} \\
        \textbf{group/type} &  & \textbf{limit} & \textbf{limit} &  & \textbf{modification\footnotemark} & \textbf{scale} \\ \hline
        Landscape & SURLAG & 1 & 12 & Surface runoff lag coefficient & Replace & HRU \\ \hline
        Landscape & CN & 40 & 95 & SCS runoff curve number & Relative & HRU \\ \hline
        Groundwater & \textbf{RCHRG\_DP} & 0 & 1 & Deep aquifer percolation fraction & Replace & HRU \\ \hline
        &  &  &  & Threshold depth of water &  &  \\ 
        Groundwater & GWQMN & 0 & 5000 & in the shallow aquifer required & Replace & HRU \\
        &  &  &  & for return flow to occur (mm) & &  \\ \hline
        Groundwater & GW\_REVAP & 0 & 0.2 & Groundwater `revap' coefficient & Replace & HRU \\ \hline
        &  &  &  & Threshold depth of water &  &  \\ 
        Groundwater & REVAPMN & 1 & 500 & in the shallow aquifer & Replace & HRU \\
        &  &  &  & for `revap' to occur (mm) & &  \\ \hline
        Groundwater & GW\_DELAY & 1 & 100 & Groundwater delay (days) & Replace & HRU \\ \hline
        Groundwater & \textbf{ALPHA\_BF} & 0.01 & 0.99 & Baseflow alpha factor & Replace & HRU \\ \hline
        Soil & SOL\_K & 0.001 & 1000 & Saturated hydraulic conductivity & Relative & HRU \\ 
        & &  & &  (mm $\mathrm{h}^{-1}$) &  &  \\ \hline
        &  &  &  & Available water &  &  \\
        Soil & SOL\_AWC & 0.01 & 0.35 & capacity of the soil layer & Relative & HRU \\
        &  &  &  & (mm $\mathrm{H}_2\mathrm{O}$ mm $\mathrm{soil}^{-1}$) &  &  \\ \hline
        Soil & ESCO & 0.01 & 1 & Soil evaporation compensation factor & Replace & HRU \\ \hline
        Soil & OV\_N & 0.008 & 0.6 & Manning's `n' value for overland flow & Replace & HRU \\ \hline
        &  &  &  & Effective hydraulic &  &  \\ 
        Channel & \textbf{CH\_K2} & 0 & 200 & conductivity in main & Replace & Sub-basin \\
        &  &  &  & channel alluvium (mm $\mathrm{h}^{-1}$) & &  \\ \hline
        Channel & \textbf{CH\_N2} & 0.016 & 0.15 & Manning's `n' value for the main channel & Replace & Sub-basin \\ \hline
        Snow & \textbf{SFTMP} & -5 & 5 & Snowfall temperature ($^{\mathrm{o}}\mathrm{C}$) & Replace & Basin \\ \hline
        Snow & \textbf{SMTMP} & -5 & 5 & Snow melt base temperature ($^{\mathrm{o}}\mathrm{C}$) & Replace & Basin \\ \hline
        & &  &  & Maximum melt rate &  &  \\ 
        Snow & \textbf{SMFMX} & 1.4 & 6.9 & for snow during the year & Replace & Basin \\ 
        & &  &  & (mm $\mathrm{H}_2\mathrm{O}$ $^{\mathrm{o}}\mathrm{C}$ $\mathrm{day}^{-1}$) &  &  \\ \hline
        & &  &  & Minimum melt rate &  &  \\ 
        Snow & SMFMN & 1.4 & 6.9 & for snow during the year & Replace & Basin \\ 
        & &  &  & (mm $\mathrm{H}_2\mathrm{O}$ $^{\mathrm{o}}\mathrm{C}$ $\mathrm{day}^{-1}$) &  &  \\ \hline
        Snow & TIMP & 0.01 & 1 & Snowpack temperature lag factor & Replace & Basin \\ \hline
        Plant & EPCO & 0.01 & 1 & Plant uptake compensation factor & Replace & Basin \\ \hline
        Plant & CANMX & 0 & 10 & Maximum canopy storage (mm $\mathrm{H}_2\mathrm{O}$) & Replace & HRU \\ \hline
    \end{tabular}
    \label{table:SWAT_Params_stats}
\end{table}
\footnotetext[1]{Table-\ref{table:SWAT_Params_stats}:~Note that the sensitive parameters are identified using the MI method.
The top seven sensitive parameters are boldfaced in this parameter column.}
\footnotetext[2]{In Table-\ref{table:SWAT_Params_stats}, the parameter modification column indicates how \texttt{SWAT} model parameters are modified during calibration and the training data generation for DL-enabled inverse modeling. 
`Replace' indicates that the existing values are replaced with values in the
provided range.
`Relative' means relative changes in \texttt{SWAT} model parameters by multiplying existing values.}

\section{DEEP LEARNING METHODOLOGY}
\label{Sec:S3_DNN_Calib}
\begin{figure}
  \centering
    \includegraphics[width = 0.95\textwidth]{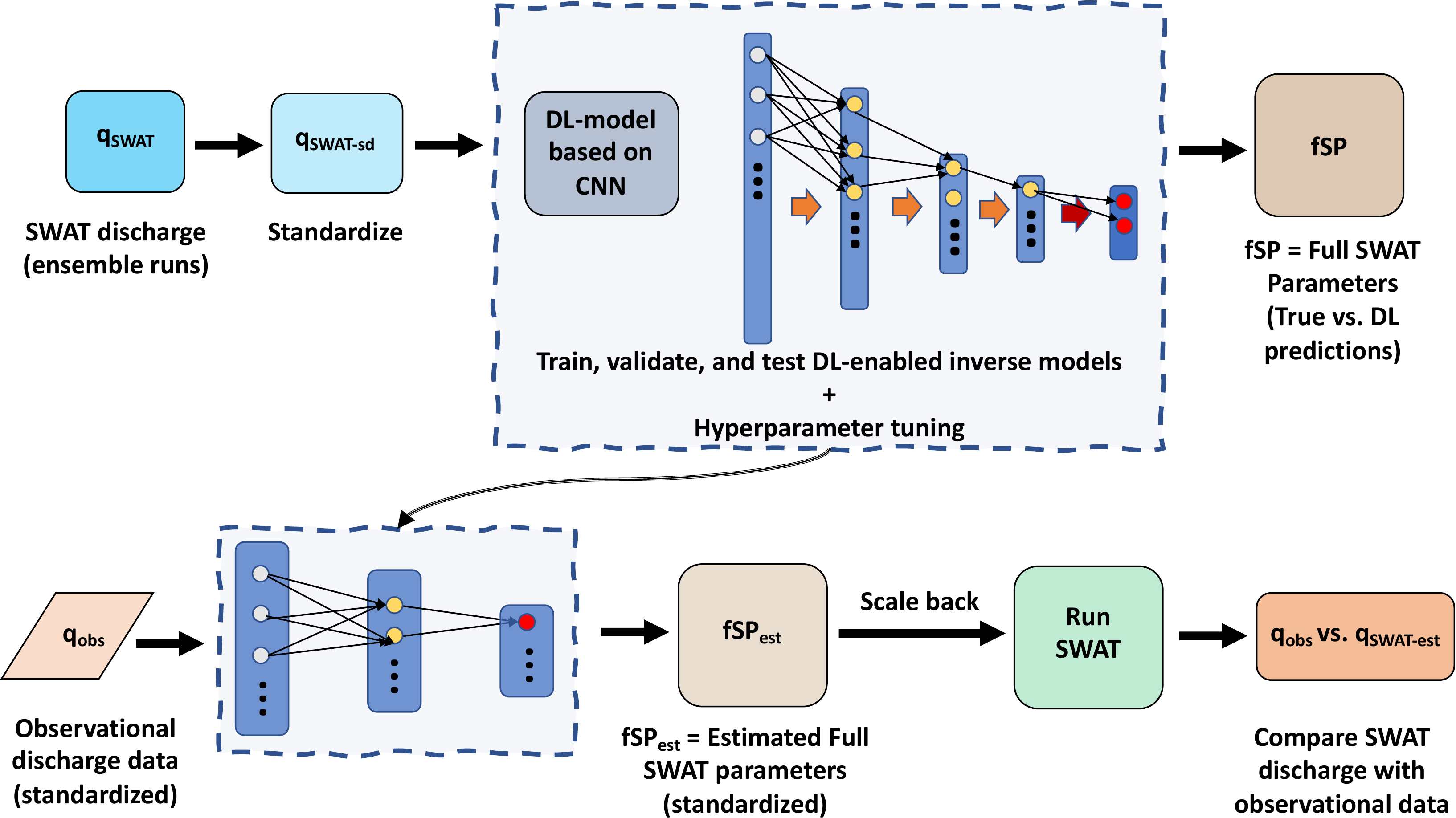}
  \caption{\textbf{Proposed deep learning workflow for the \texttt{SWAT} model calibration:}~A pictorial description of the proposed DL workflow to estimate parameters and calibrate the \texttt{SWAT} model using observational discharge.
  Ensemble simulations generated by the \texttt{SWAT} model are used to train, validate, and test the DL-enabled inverse models.
  The observational data is then provided as an input to the developed DL models to estimate site-specific parameters.
  These parameters are then used by the \texttt{SWAT} model to simulate discharge for comparison with observational data.
  \label{Fig:DL_Workflow_SWAT}}
\end{figure}

This section presents the overall methodology consisting of data pre-processing, hyperparameter tuning, and the computational cost of constructing the DL-enabled inverse models.
Figure~\ref{Fig:DL_Workflow_SWAT} shows the DL workflow for
training the inverse model and then inferring the \texttt{SWAT} parameters. 
The DL workflow can be divided into two main steps.
In the first step, we train, validate, and test DL-enabled inverse models based on convolutional neural networks (CNNs) \cite{schmidhuber2015deep,Goodfellow-et-al-2016,chollet2017deep} using \texttt{SWAT} model ensemble runs.
The synthetic streamflow and parameter sets are assembled into a data matrix and then partitioned into training (80\%), validation (10\%), and testing (10\%) sets.
Each \texttt{SWAT} run contains 3654 daily measurements, which are normalized by removing the mean and scaling the training dataset to unit variance. 
This standardizes the raw datasets into a representation suitable for training CNNs.
The resulting pre-processing estimator is then used to transform the validation and testing sets. 
Pre-processing is necessary for DL model development, as CNNs are filter/kernel-based methods that benefit from the normalization of their inputs to make accurate predictions \cite{anysz2016influence,gu2018recent}.
Hyperparameter tuning is performed to identify the optimal CNN architectures.
In the second DL workflow step, the observational data is standardized using the pre-processing estimator.
This normalized data is input to the tuned DL-enabled inverse models to estimate the study site parameters.
These calibrated parameter sets are given to the \texttt{SWAT} model to obtain daily streamflow values in the calibration and validation periods.
The predicted discharge is then compared with the raw observational data to measure the performance of the DL-enabled calibrated \texttt{SWAT} model.

\subsection{CNN architectures based on hyperparameter tuning}
\label{SubSec:S3_hp_tuning}
Three different types of DL-enabled inverse models are developed based on CNNs.
First, a CNN architecture is used as multi-task learning (MTL) model, which estimates all the 21 parameters.
Correlations between the \texttt{SWAT} parameters are accounted for in this 21-parameter model. 
The second architecture is also an MTL model but focused on estimating the top seven sensitive parameters.
This 7-parameter model estimates the boldfaced parameters listed in Table~\ref{table:SWAT_Params_stats}.
The third architecture is based on a single-task learning (STL) model, which estimates one sensitive \texttt{SWAT} parameter.
We, therefore, trained seven different 1-parameter models to estimate all the sensitive parameters. 

Hyperparameter tuning is performed to identify the optimal CNN architecture of the 21-, 7-, and 1-parameter models.
This tuning is necessary as the training process, and predictions of the DL-enabled inverse model are controlled by the CNN architecture's parameters and topology.
We tested two types of hyperparameters:~(1) model hyperparameters and (2) algorithm hyperparameters.
Model hyperparameters define the neural network architecture.
For instance, the selection of CNN topology is influenced by model hyperparameters such as the number and width of hidden layers.
Algorithm hyperparameters influence the training process after the architecture is established.
The values of the trainable weights of a CNN architecture are controlled by algorithm hyperparameters such as learning rate and the number of epochs.
Table~\ref{table:Hyperparameter_21_7_1} shows the search space and best set of both model and algorithm hyperparameters.
During the tuning process, the batch size is kept constant (equal to 10) and ReLU is the activation function. 
The optimal hyperparameter set is chosen based on the validation mean squared error using the grid search tuning method.
In addition to identifying the optimal hyperparameter set, we also identified the next best candidates using validation mean squared error.
Table S1 in the supplementary material shows the hyperparameters of the remaining nine best DL-enabled inverse models. 
Note that we obtained a total of ten best hyperparameter sets for each of the 21-, 7-, and 1-parameter model scenarios.
In Sec.~\ref{Sec:S4_Results}, we show the predictions of these ten best models and the associated uncertainty in their streamflow predictions.

\begin{figure}
  \centering
    \includegraphics[width = 0.95\textwidth]{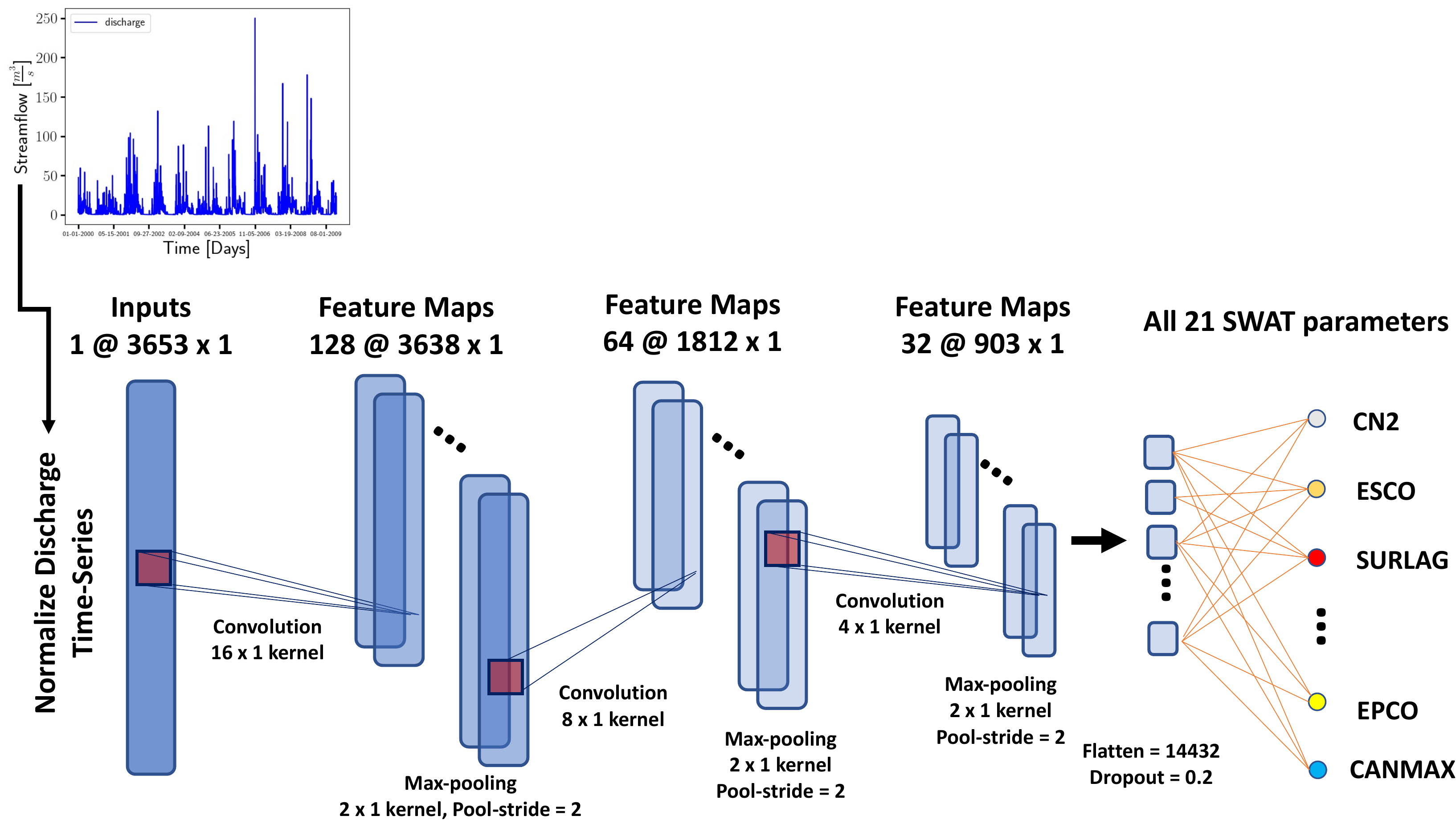}
  \caption{\textbf{A tuned DL-enabled inverse model architecture:}~This figure shows a pictorial description of a deep convolutional neural network for estimating \texttt{SWAT} model parameters.
  This DL model inversely maps simulated discharge to 21 different conceptual parameters.
  Hyperparameter tuning is then performed to arrive at the CNN architecture.
  This DL-enabled inverse model is a multi-task learning model, which accounts for the correlation between different \texttt{SWAT} parameters during training.
  \label{Fig:DL_Baseline_Model}}
\end{figure}

We used the \textsf{Keras} API in \textsf{Tensorflow} package \cite{tf-keras} to build our DL-enabled inverse models.
Figure~\ref{Fig:DL_Baseline_Model} shows a pictorial description of the tuned architecture of the 21-parameter model.
Figures S2 and S3 in the supplementary material show pictorial descriptions of the tuned architectures for the 7- and 1-parameter models, respectively.
The CNN filters are initialized with the Glorot uniform
initializer.
After each convolution, a max-pooling operation is applied and the final convolutional layer is then flattened.
After the dropout layer, the remaining features are mapped to the \texttt{SWAT} parameters.
The entire CNN is compiled using an Adam optimizer, with the loss being the mean squared error.
The resulting tuned 21-, 7-, and 1-parameter models have a total of 379,093; 177,031; 90,433 trainable weights, respectively.

\begin{table}[htbp]
    \centering
    \caption{This table provides the hyperparameter space to develop and tune 21-, 7- and 1-parameter models.
    The optimal set identified is boldfaced in the `Explored options' column.
    The tuned CNN architecture of the 21-parameter model is shown in Fig.~\ref{Fig:DL_Baseline_Model}.}
    \scriptsize
    \begin{tabular}{|c|c|c|}\hline
        \textbf{Hyperparameter type} & \textbf{Description} & \textbf{Explored options} \\ \hline
        Layers & Number of 1D convolutional layers & $[1, 2, \mathbf{3}, 4, 5]$ \\ \hline
        Filters & The number of output filters in the 1D convolution & $[16, \mathbf{32, 64, 128}, 256]$ \\ \hline
        Kernel size & An integer to specify the length of the 1D convolution window & $[2, \mathbf{4, 8, 16}, 32]$ \\ \hline
        Dropout rate & Applies dropout to the input\footnotemark & $[0.0, 0.1, \mathbf{0.2}, 0.3, 0.4]$ \\ \hline
       Learning rate & The value of the optimizer in the Adam algorithm & $[10^{-6}, \mathbf{10^{-5}}, 10^{-4}]$ \\ \hline
       Epochs & The number of times the algorithm sees the training data & $[100, 200, 300, 400, \textbf{500}, 1000]$ \\ \hline
    \end{tabular}
    \label{table:Hyperparameter_21_7_1}
\end{table}
\footnotetext[3]{Table-\ref{table:Hyperparameter_21_7_1}:~To reduce model overfitting, we randomly set the last convolutional layer units that connect to the output to 0 at each step during training time.
The rate value controls the frequency of dropping the units.}

\subsection{Computational cost}
\label{SubSec:S3_Comp_Cost}
The wall clock time to run a single ten-year \texttt{SWAT} model simulation is approximately 240 seconds on a four-core processor (Intel(R) i7-8650U CPU @ 1.90GHz), a standard desktop machine.
The ensemble run simulations for training the DL-enabled inverse models were developed using a cluster of 56 cores (Intel(R) Xeon(R) Gold 5120 CPU @ 2.20GHz) and 256 GB DDR4 RAM.
We trained the proposed models on a MacBook Pro Laptop (2.3GHz 8-Core Intel i9 CPU, 64GB DDR4 RAM) with an associated computational cost of 5 seconds per epoch per model.
The wall clock time needed to generate a best tuned DL-enabled inverse model is approximately 2500 seconds, which can be accelerated using GPUs.
After the DL model is trained, the computational inference cost to estimate the \texttt{SWAT} model parameters is 0.16 seconds.
The wall clock time for making a prediction/inference shows that our DL-enabled parameter estimation is at least $\mathcal{O}(10^3)$ times faster than the GLUE-based method (e.g., may require 100-1000s of forward model runs), in addition to its predictive capability. 
Our future work involves accelerating the training process using GPUs available at leadership class supercomputing resources (e.g., NERSC, OLCF, and ALCF user facilities) \cite{ALCF2021,OLCF2021,NERSC2021}.

\section{RESULTS AND DISCUSSION}
\label{Sec:S4_Results}
This section presents results on the overall accuracy and efficiency of the proposed DL methodology. 
First, we describe the DL-enabled inverse modeling results from the ensemble runs.
Second, we show the \texttt{SWAT} model parameters estimated from observational discharge and compare the performance of DL-enabled parameter estimation and the GLUE method.
We then compare the streamflow predictions from the calibrated \texttt{SWAT} model with the observation discharge for both the calibration and validation periods.
Next, we provide the sensitivities of the estimated \texttt{SWAT} model parameters under a range of relative errors.
Finally, we give the performance metrics, and calibration uncertainties for DL-enabled and GLUE estimated parameters.

\begin{figure}
  \centering
    \subfigure[21-parameter model:~Loss]
    {\includegraphics[width = 0.295\textwidth]{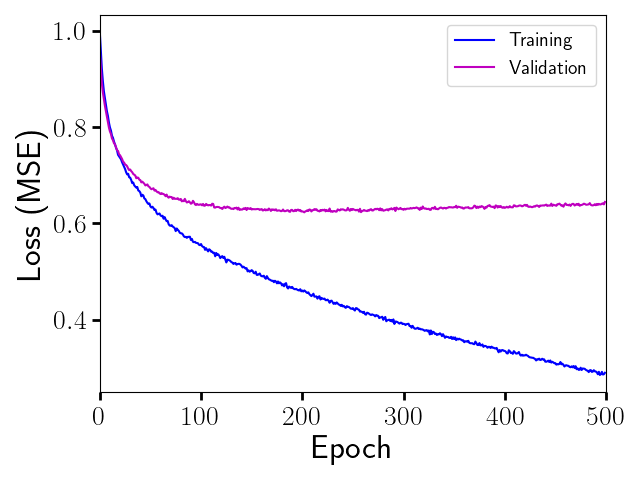}}
    \subfigure[7-parameter model:~Loss]
    {\includegraphics[width = 0.295\textwidth]{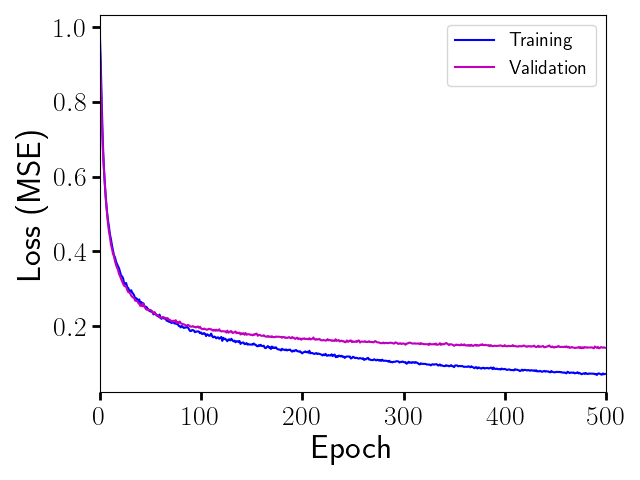}}
    \subfigure[1-parameter model:~Loss (SFTMP)]
    {\includegraphics[width = 0.295\textwidth]{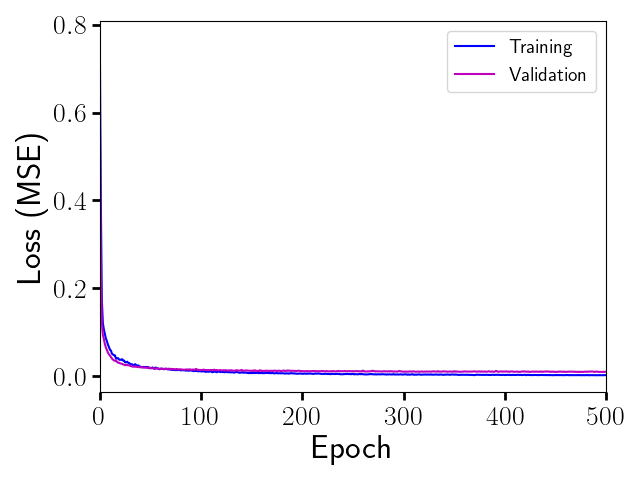}}
  \caption{\textbf{Loss metrics of best inverse models:}~This figure shows the overall training and validation loss of three different DL-enabled inverse models.
  The overall loss of the 21-parameter model is the sum of the loss of individual \texttt{SWAT} parameters.
  Similarly, the overall loss of the 7-parameter model is the sum of the loss of the top seven sensitive \texttt{SWAT} parameters.
  The loss of the 1-parameter model corresponds to the SFTMP, which is the top sensitive parameter.
  \label{Fig:DL_Models_Loss_SFTMP}}
\end{figure}

\subsection{Training, validation, and testing results}
\label{SubSec:S4_TrainValTest_Results}
Figure~\ref{Fig:DL_Models_Loss_SFTMP} shows the training and validation loss of the best DL-enabled inverse models in estimating the \texttt{SWAT} parameters.
The MSE values in Fig.~\ref{Fig:DL_Models_Loss_SFTMP}(a) show the overall loss for all 21 \texttt{SWAT} parameters.
Based on the tuned 7-parameter model, Fig.~\ref{Fig:DL_Models_Loss_SFTMP}(b) provides the overall loss of the top seven sensitive parameters.
Figure~\ref{Fig:DL_Models_Loss_SFTMP}(c) shows the loss in estimating SFTMP based on the 1-parameter model developed for that most sensitive parameter.
The supplementary figure S4 shows the loss for the other six 1-parameter models that estimate the remaining sensitive parameters.
Figure S4(g) also compares the sum of losses of the 1-parameter models for the top seven sensitive parameters with that of the 7-parameter model.
The 7-parameter model's losses are lower than the sum of 1-parameter models as the 7-parameter model's training accounts for the correlation between sensitive parameters, producing a better MSE value than the sum of 1-parameter models.
Furthermore, these figures show that all the tuned models have converged.
The validation loss of the 21-parameter model plateaus after 200 epochs, even as the training loss decreases.
Over-fitting is due to the lack of valuable information in the streamflow data to constrain the lesser and insensitive parameters.
Over-fitting is less pronounced in the 7-parameter model as the validation loss is much closer to training.
The 1-parameter model estimating SFTMP does not demonstrate this over-fitting problem.
Similarly, the 1-parameter models for ALPHA\_BF, CH\_K2, and SMTMP sensitive parameters do not show over-fitting problems (see Fig.~S4).
However, over-fitting occurs for RCHRG\_DP and, to a lesser extent, the CH\_N2 and SMFMX parameters as their validation losses closely follow the training losses.
As the sensitivity of the parameters to discharge decreases, over-fitting occurs.

\begin{figure}
  \centering
    \subfigure[21-parameter model:~Training]
    {\includegraphics[width = 0.295\textwidth]{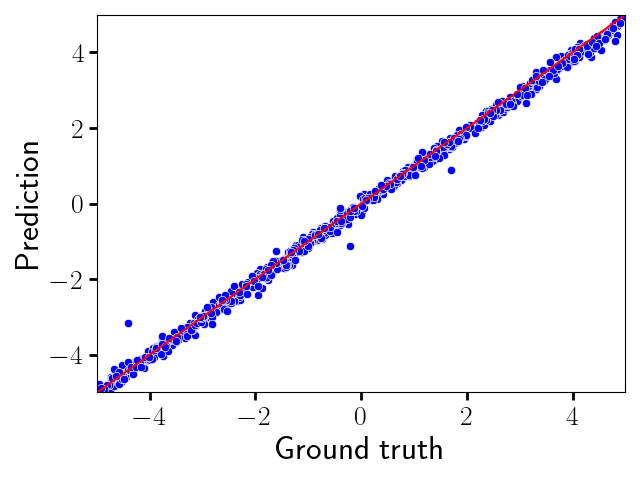}}
    \subfigure[21-parameter model:~Validation]
    {\includegraphics[width = 0.295\textwidth]{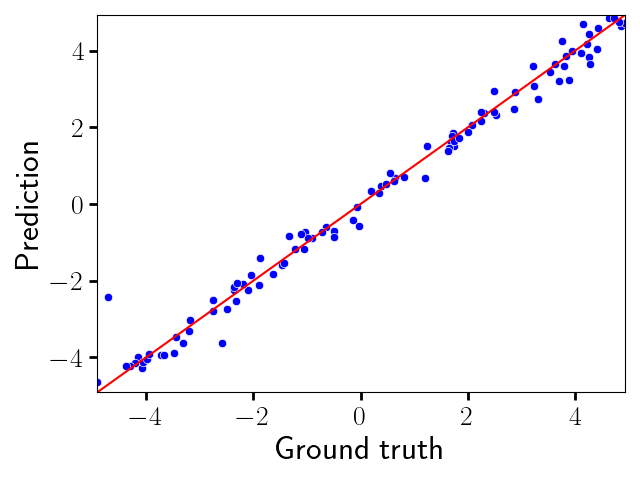}}
    \subfigure[21-parameter model:~Testing]
    {\includegraphics[width = 0.295\textwidth]{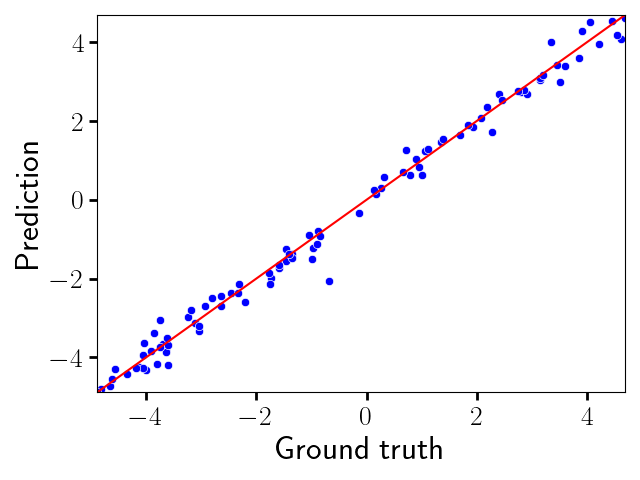}}
    \subfigure[7-parameter model:~Training]
    {\includegraphics[width = 0.295\textwidth]{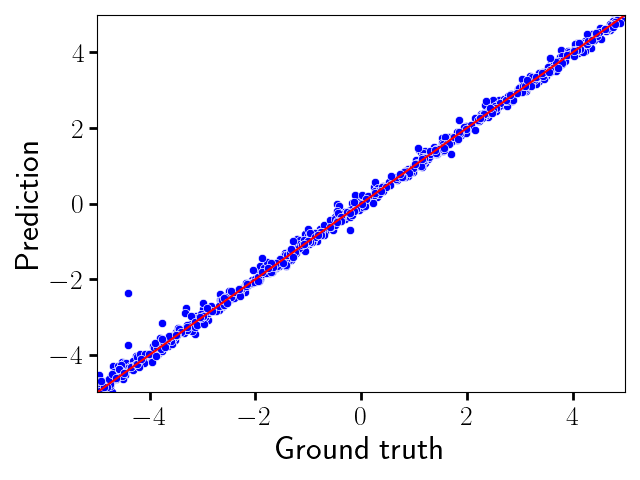}}
    \subfigure[7-parameter model:~Validation]
    {\includegraphics[width = 0.295\textwidth]{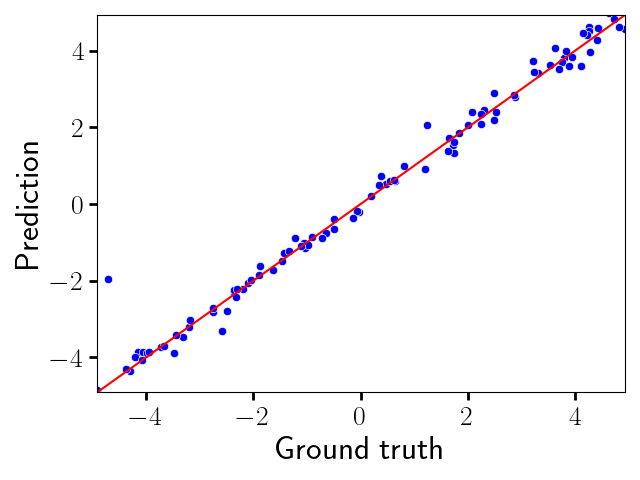}}
    \subfigure[7-parameter model:~Testing]
    {\includegraphics[width = 0.295\textwidth]{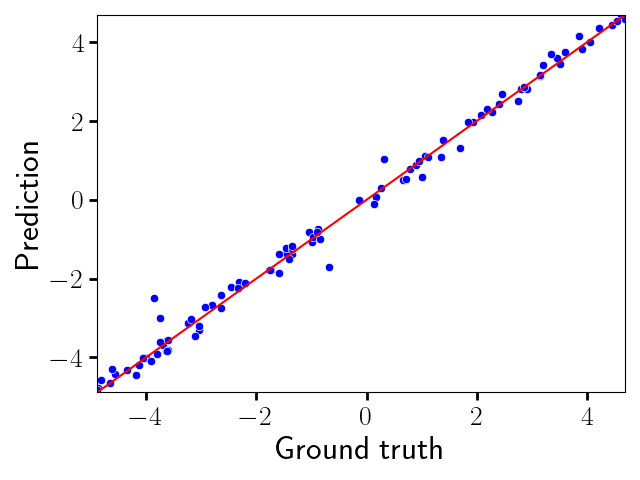}}
    \subfigure[1-parameter model:~Training]
    {\includegraphics[width = 0.295\textwidth]{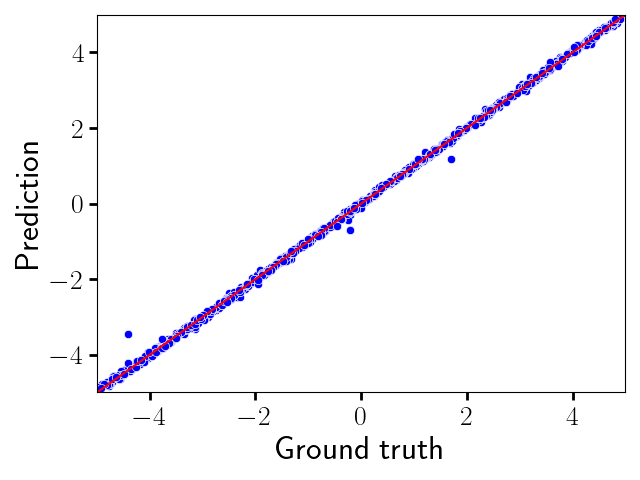}}
    \subfigure[1-parameter model:~Validation]
    {\includegraphics[width = 0.295\textwidth]{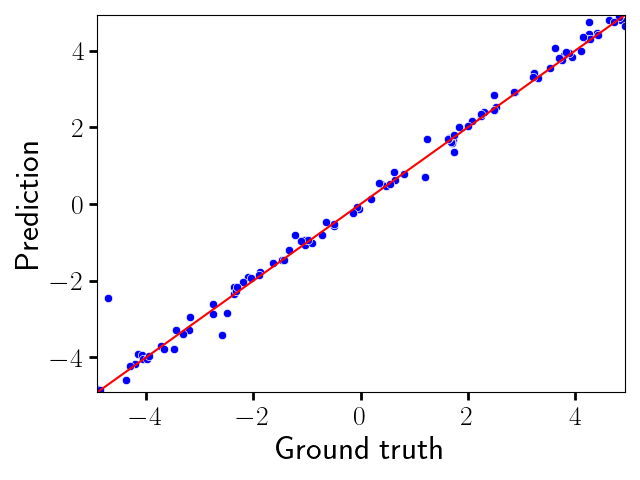}}
    \subfigure[1-parameter model:~Testing]
    {\includegraphics[width = 0.295\textwidth]{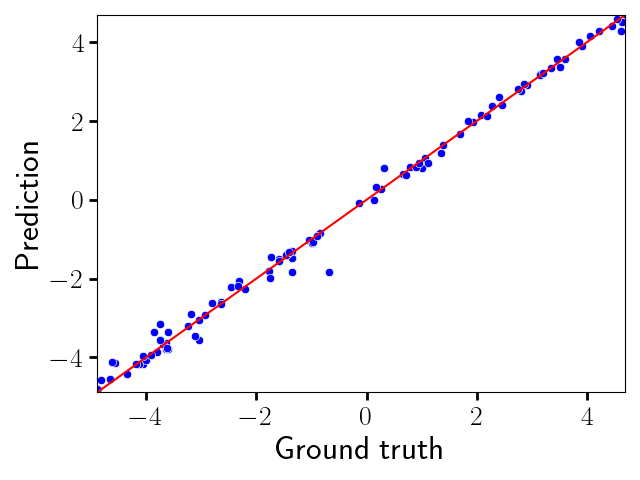}}
  \caption{\textbf{Predictions of DL-enabled inverse models for the SFTMP:}~This figure shows one-to-one plots of the best inverse models for the most sensitive parameter, SFTMP (units in $^{\mathrm{o}}\mathrm{C}$).
  It compares the DL estimation with the ground truth for the training, validation, and test datasets. 
  Each blue dot represents a realization from the corresponding train/validation/test set of ensembles. 
  The red line is the one-to-one line.
  \label{Fig:DL_Models_SFTMP_Predictions}}
\end{figure}

Figure~\ref{Fig:DL_Models_SFTMP_Predictions} shows the prediction of the tuned DL-enabled inverse models for estimating SFTMP.
Supplementary figures S5-S10 provide the predictions of the best models for the remaining sensitive parameters.
All three models estimate the most sensitive parameters (SFTMP, ALPHA\_BF, SMTMP, CH\_N2, and CH\_K2) well using the training set.
The one-to-one plots between the estimated and true parameters are closely distributed along the 1:1 line.
The accuracy of the training predictions is lower for SMFMX and RCHRG\_DP.
This reduced accuracy is evident from the more scattered drift away from the one-to-one straight line, seen in Fig.~S5 and S10 in the supplementary material.
This indicates that the CNNs could not extract sufficient information from the discharge time series to estimate these less sensitive parameters.

\begin{figure}
  \centering
    \subfigure[1- vs. 21-parameter model]
    {\includegraphics[width = 0.295\textwidth]{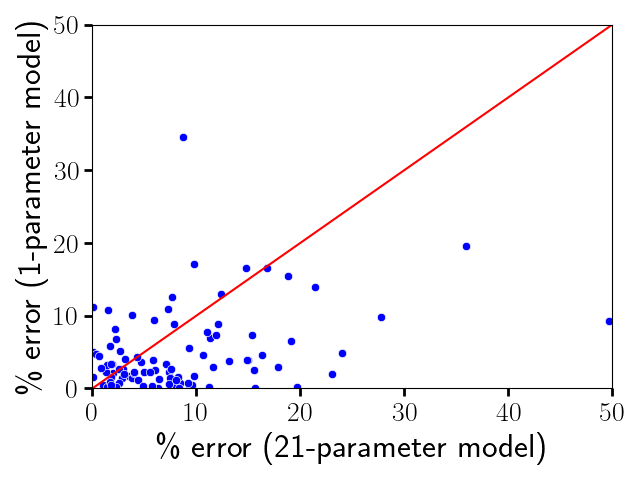}}
    \subfigure[7- vs. 21-parameter model]
    {\includegraphics[width = 0.295\textwidth]{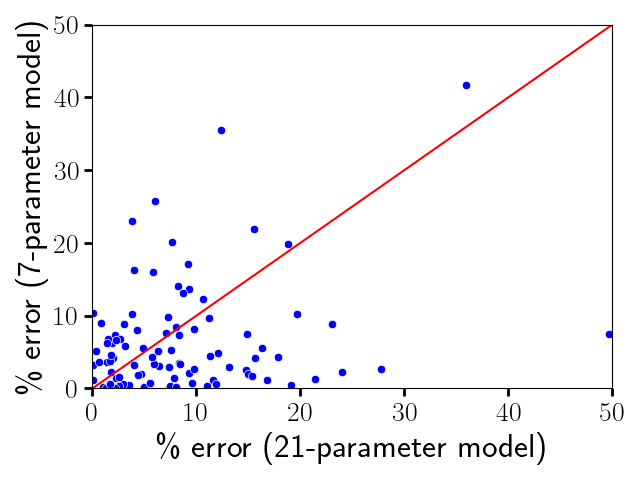}}
    \subfigure[1- vs. 7-parameter model]
    {\includegraphics[width = 0.295\textwidth]{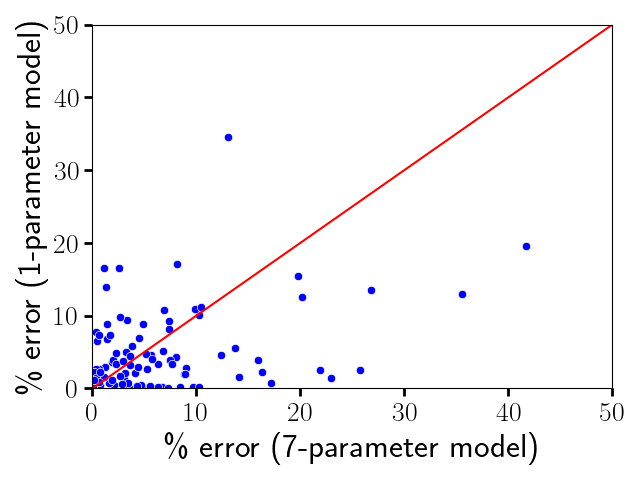}}
  \caption{\textbf{Comparison of the best DL-enabled inverse models in predicting SFTMP on the test dataset:}~This figure provides a one-to-one plot of percent errors in predicting SFTMP.
  The DL analysis is shown for the test data, where each data point represents a realization from this set.
  The red line is the one-to-one line.
  The DL models (i.e., 1-parameter and 7-parameter models) guided by sensitivity analysis slightly outperform the 21-parameter models.
  This is because the percent errors of such models are less than those of the 21-parameter model.
  \label{Fig:DL_SFTMFP_PercentError_Testdata}}
\end{figure}

For the validation and testing sets, the tuned 21-, 7-, and 1-parameter models perform very well for estimating SFTMP.
This is evident from the one-to-one scatter plots shown in Fig.~\ref{Fig:DL_Models_SFTMP_Predictions}.
Similar inferences can be drawn for the parameters ALPHA\_BF, CH\_N2, and CH\_K2, although they have lesser estimation accuracy.
Figs.~S5, S9, and S10 demonstrate the disparity in estimation accuracies for RCHRG\_DP, SMTMP, and SMFMX, respectively.
This reduced accuracy is comparable to the training results, where we see an increased deviation of the scatters from the one-to-one straight line.
Figure~\ref{Fig:DL_SFTMFP_PercentError_Testdata} shows the performance of all the three tuned DL-enabled inverse models on the testing set for SFTMP.
The supplementary Figs.~S11 to S16 compare the performance of the other six sensitive parameters.
These figures show that all three models provide very similar results in terms of percent errors.
For specific scenarios (e.g., SFTMP), the 1- and 7-parameter models provide slightly better estimates than the 21-parameter model.
This can be attributed to contributions of the less sensitive parameters that potentially reduce the performance of the 21-parameter model.
Figures S29-S35 show the performance of the remaining nine best DL-enabled inverse models on test datasets.
These tuned architectures are identified using validation mean squared error and produce similar predictions to those of the best DL-enabled inverse model.
These other models can accurately estimate the most sensitive parameter, SFTMP, but have reduced accuracy for less sensitive parameters.

\subsection{Sensitivity of estimated \texttt{SWAT} parameters to observation noise}
\label{SubSec:S4_Test_Discharge_Errors}
We selected all test realizations to evaluate the parameter estimation sensitivity of the DL-enabled inverse models to observational errors.
We added random observation errors to the synthetic observed discharge time series for each test realization.
We then generated 100 different observation realizations for parameter estimation, i.e., ($\mathbf{q}_n$), which is given by 
\begin{equation}
  \label{Eqn:Obs_Error}
  \mathbf{q}_n = \mathbf{q} + \epsilon \times \mathbf{q} \times \mathbf{r}
\end{equation}
where $\epsilon$ is the standard deviation of the noise, usually taken as $\frac{1}{3}$ of the observation error, and $\mathbf{r}$ is a random vector of the same size as $\mathbf{q}$.
The elements of the random vector contain samples drawn from a standard normal distribution with a mean of 0 and a standard deviation of 1.
We tested different levels of observation errors, i.e., 5\%, 10\%, 15\%, 20\%, and 25\% relative to the observed values.
These noisy discharge data (both synthetic and observations) are provided as input to the best DL-enabled inverse models to estimate the \texttt{SWAT} model parameters.  

\begin{figure}
  \centering
    %
    \subfigure[Noise added to a randomly selected test realization]
    {\includegraphics[width = 0.485\textwidth]{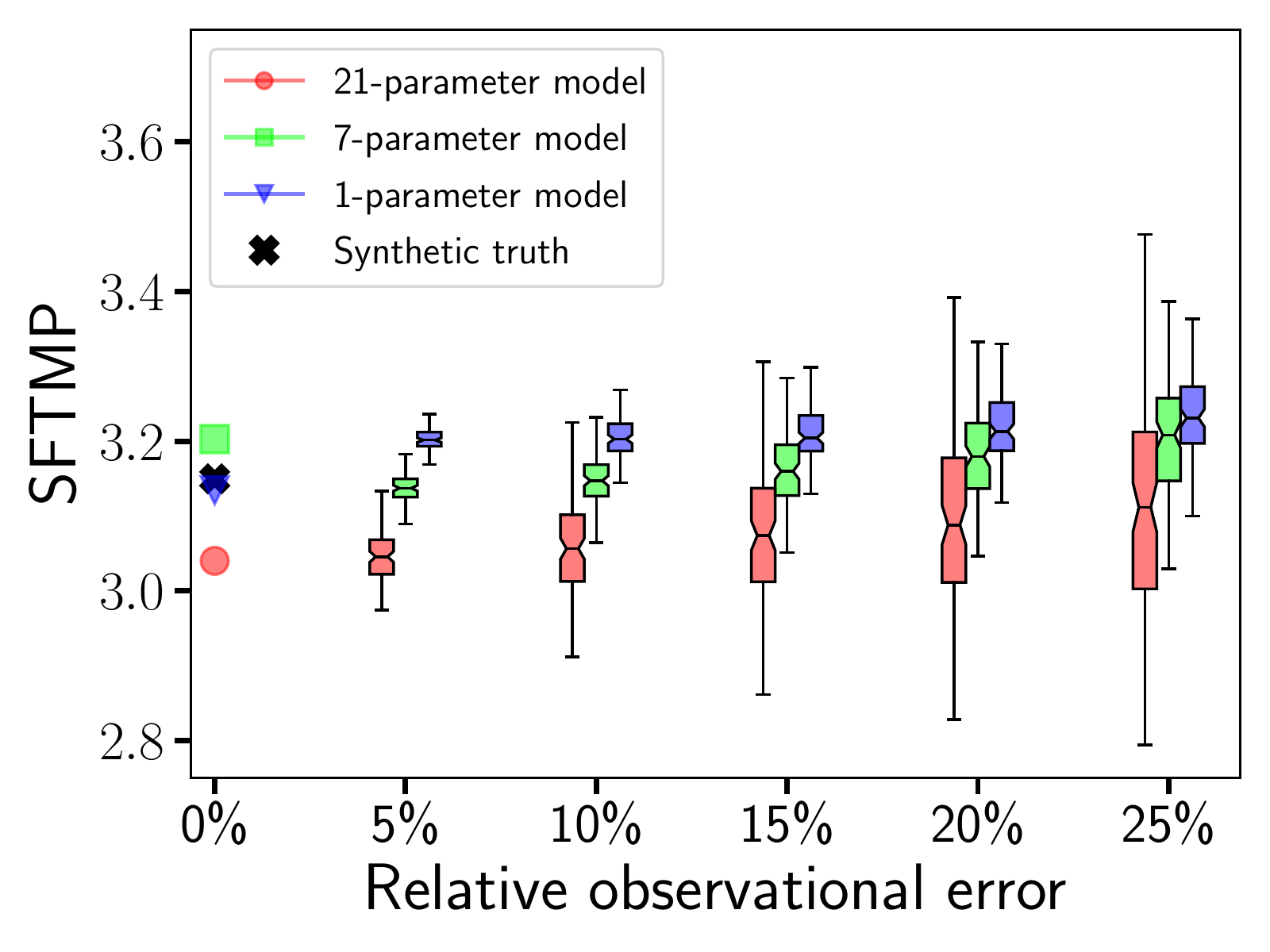}}
    \subfigure[Noise added to observational discharge (calibration period)]
    {\includegraphics[width = 0.485\textwidth]{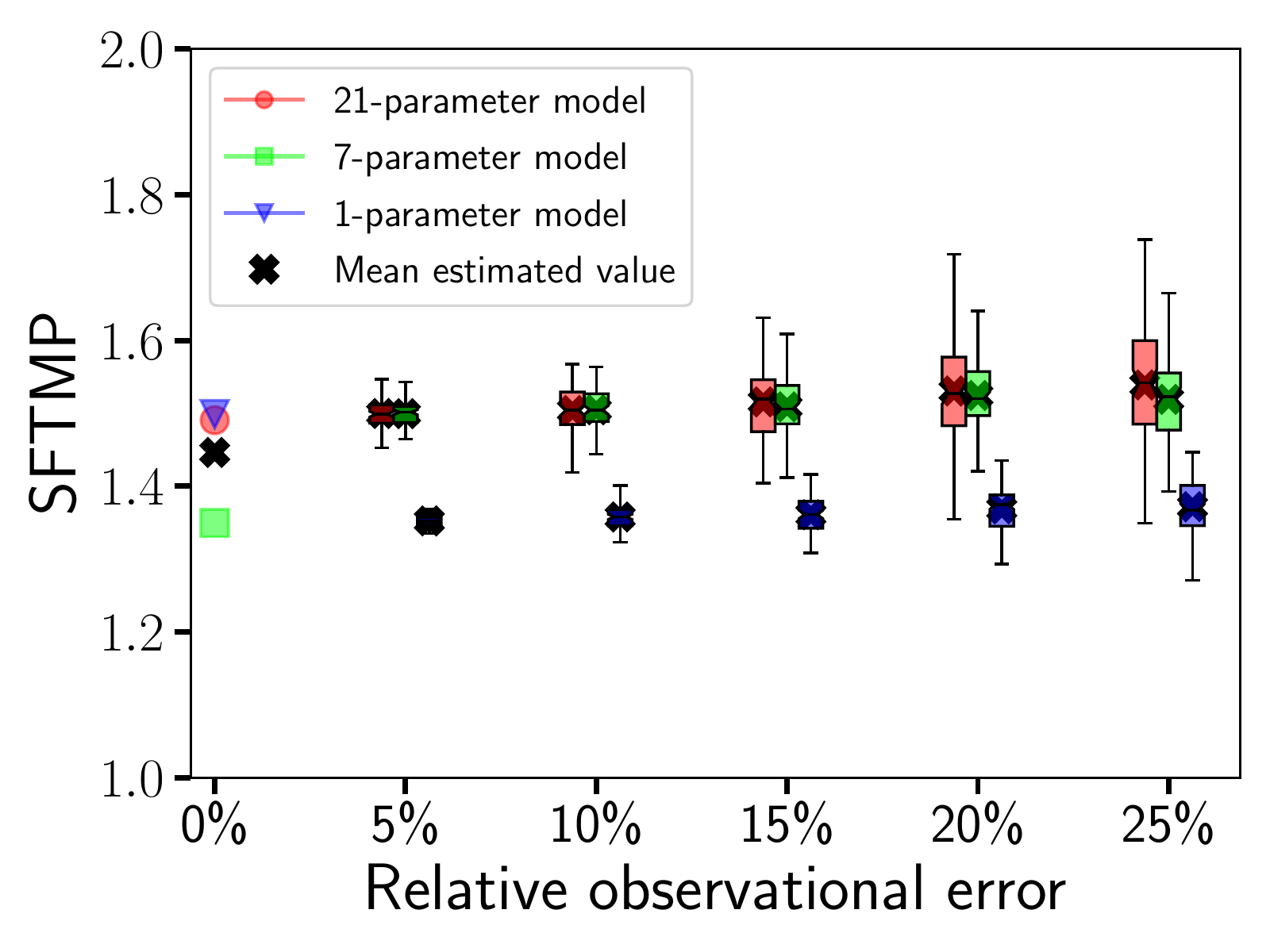}}
    \subfigure[Noise added to all test realizations]
    {\includegraphics[width = 0.985\textwidth]{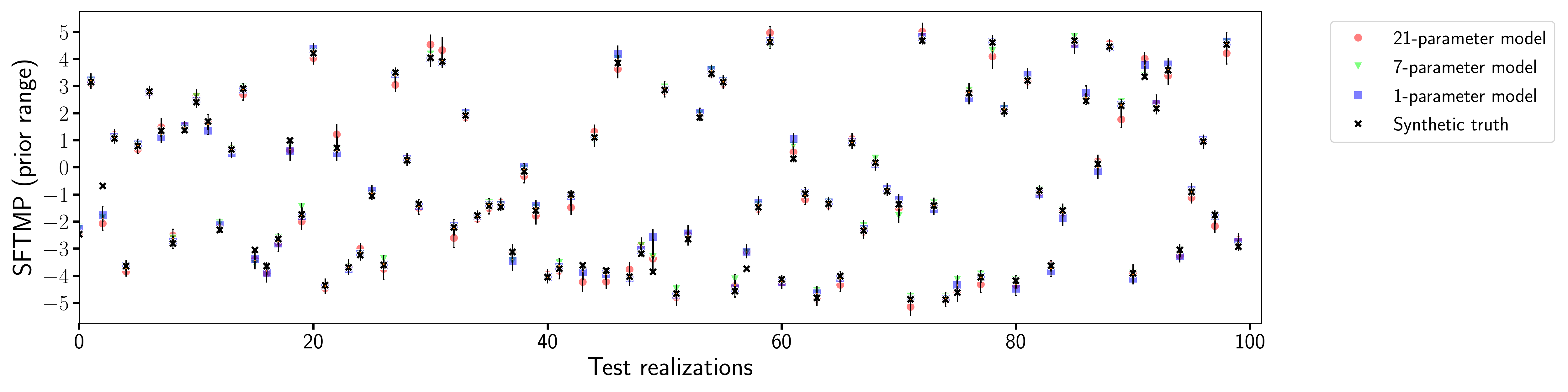}}
  \caption{\textbf{Sensitivity of DL estimated parameters to noise (SFTMP):}~This figure shows the sensitivities of the DL-enabled inverse models to different noise levels added to the simulated and observational discharge.
  The results are based on the best models represented by the colors filling the box plots.
  The top left figure shows the variation in SFTMP (units in $^{\mathrm{o}}\mathrm{C}$) with noise level for a simulated test realization.
  The markers in the top left figure show the synthetic truth and compare it with predictions under no noise.
  Similarly, the markers in the top right figure show the mean estimated value from the three best models for real data.
  The right figure also compares the estimations with different noise values.
  The maximum possible variation in SFTMP is less than 0.5$^{\mathrm{o}}\mathrm{C}$ even under 25\% relative observational errors.
  Moreover, the estimations of the most sensitive parameter SFTMP are robust to noise as the predictions are closer to ground truth for all synthetic predictions, shown in the bottom figure.
  The prediction uncertainty in SFTMP, shown by a black line, is obtained by combining the noisy predictions from 1-, 7-, and 21-parameter models.
  The red, green, and blue markers represent the mean predictions across 100 noisy samples generated for each realization.
  \label{Fig:Params_SenstoNoise}}
\end{figure}

Figure~\ref{Fig:Params_SenstoNoise} shows the variability in estimated SFTMP from the 21-, 7-, and 1-parameter models as boxplots.
It also shows DL model predictions for all noisy test realizations (see Fig.~\ref{Fig:Params_SenstoNoise}(c)).
Figs.~S23 to S28 in the supplementary material provide estimations for the other six sensitive parameters.
Note that the parameter estimations are within the prior sampling range even after adding high relative noise, which instills confidence in the predictive capabilities of DL models.
From Fig.~\ref{Fig:Params_SenstoNoise}, it is evident that all the three DL-enabled inverse models are robust to noise in estimating the most sensitive parameter.
Variation in SFTMP from the DL models is less than 0.5$^{\mathrm{o}}\mathrm{C}$ even under high observational noise (e.g., 25\%) for both test and real data.
This shows that the SFTMP predictions are not sensitive to noise, as the CNNs' performance is stable even after adding noise to data.
This predictive capability under noise also provides the insight that CNNs can effectively learn the underlying representations in the streamflow data rather than modeling noise.
Similar assessments can be made for the ALPHA\_BF, CH\_N2, RCHRG\_DP, and SMTMP parameters.
However, the predictions of the 21- and 7-parameter models for CH\_K2 seem to be influenced by noise.
The 1-parameter model seems to be more robust to noise for estimating CH\_K2 in this case.
The performance of DL estimation for SMFMX, which is the least sensitive parameter, is lower than that of sensitive parameters such as SFTMP.
As discussed in Sec.~\ref{SubSec:S4_TrainValTest_Results} and from MI analysis, it is evident that estimations are inadequate for this parameter.
This reduced performance is because less valuable information is available in the discharge data to estimate SMFMX robustly, limiting the proposed models to extract useful representation from input data to calibrate lesser sensitive parameters better, thus reducing the estimation accuracy.

\subsection{Calibrated \texttt{SWAT} model based on real observed discharge}
\label{SubSec:S4_SWAT_Calib_ObsDischarge}
The trained DL-enabled inverse models are used to estimate parameters at the ARW study site based on actual discharge observations.
As the 21-, 7-, and 1-parameter models have similar performances (shown in Subsecs.~\ref{SubSec:S4_TrainValTest_Results} and \ref{SubSec:S4_Test_Discharge_Errors}), we provide the streamflow predictions of the calibrated \texttt{SWAT} model based on 21-parameter models.
In addition to the tuned 21-parameter model, we also show the estimations of nine following best candidates.
Similar inferences can be drawn for 7- and 1-parameter models.

\begin{figure}
  \centering
    \subfigure[KGE vs. SFTMP]
    {\includegraphics[width = 0.475\textwidth]{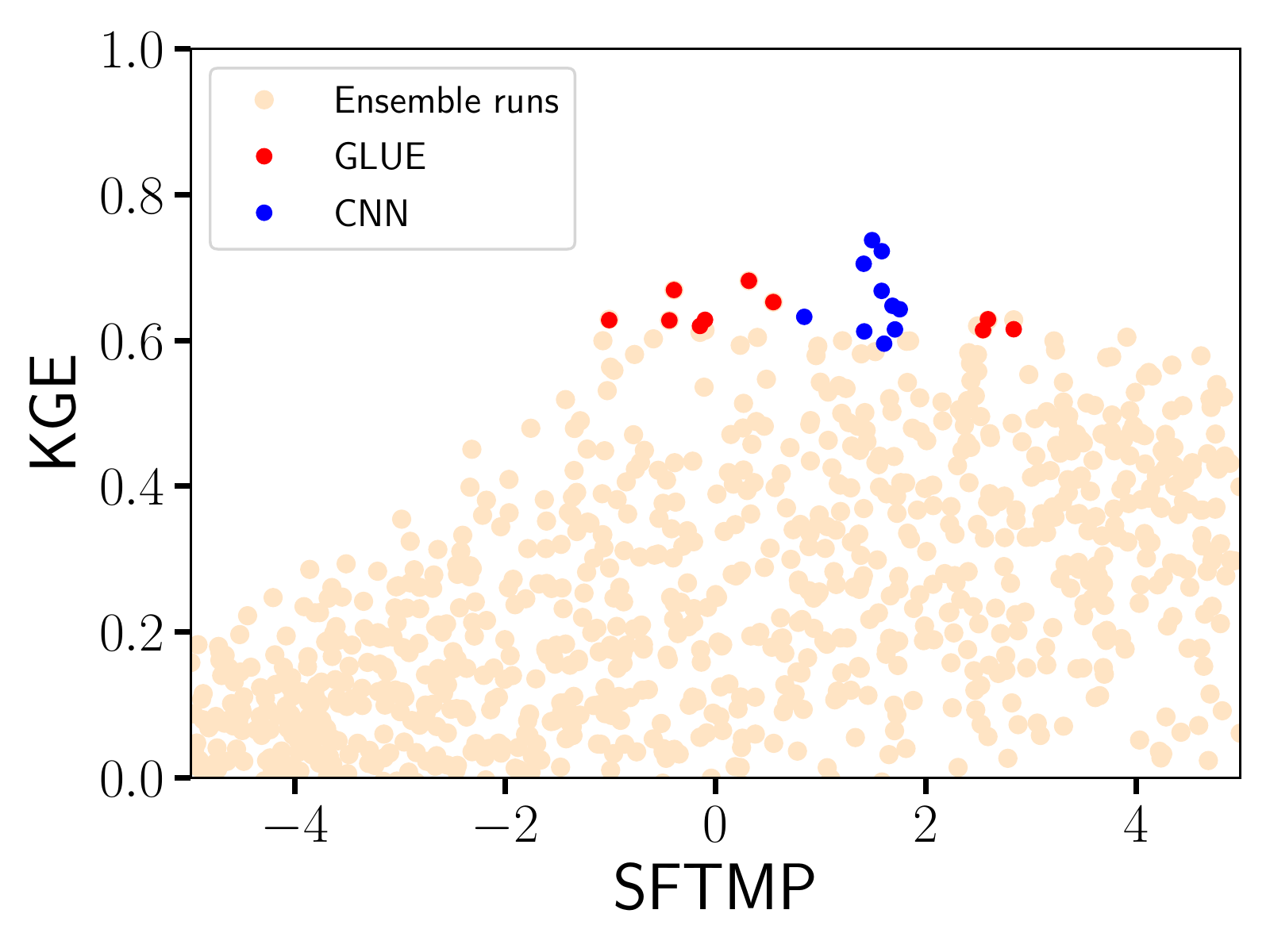}}
    \subfigure[$R^2$-score vs. SFTMP]
    {\includegraphics[width = 0.475\textwidth]{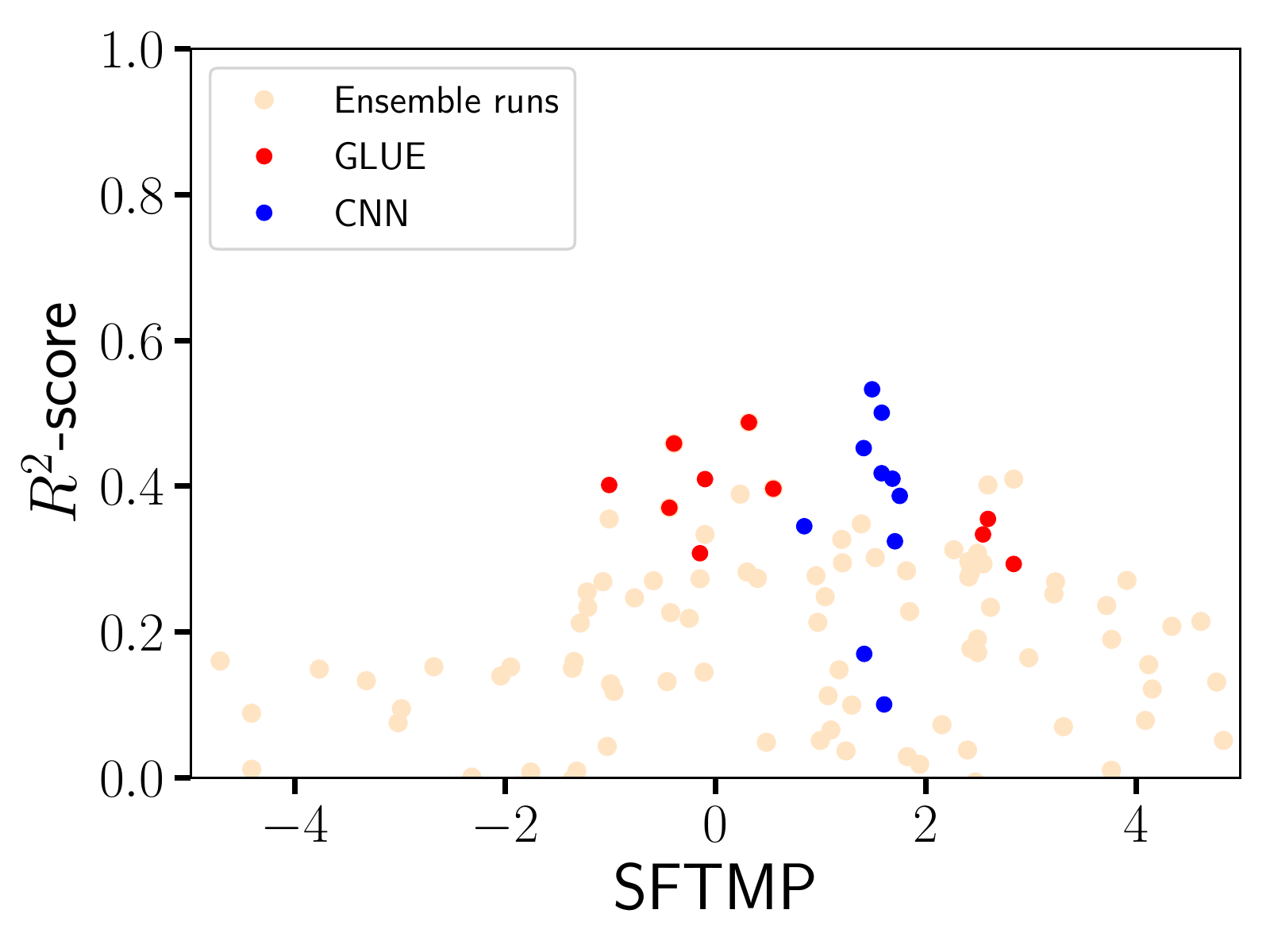}}
    \subfigure[NSE vs. SFTMP]
    {\includegraphics[width = 0.475\textwidth]{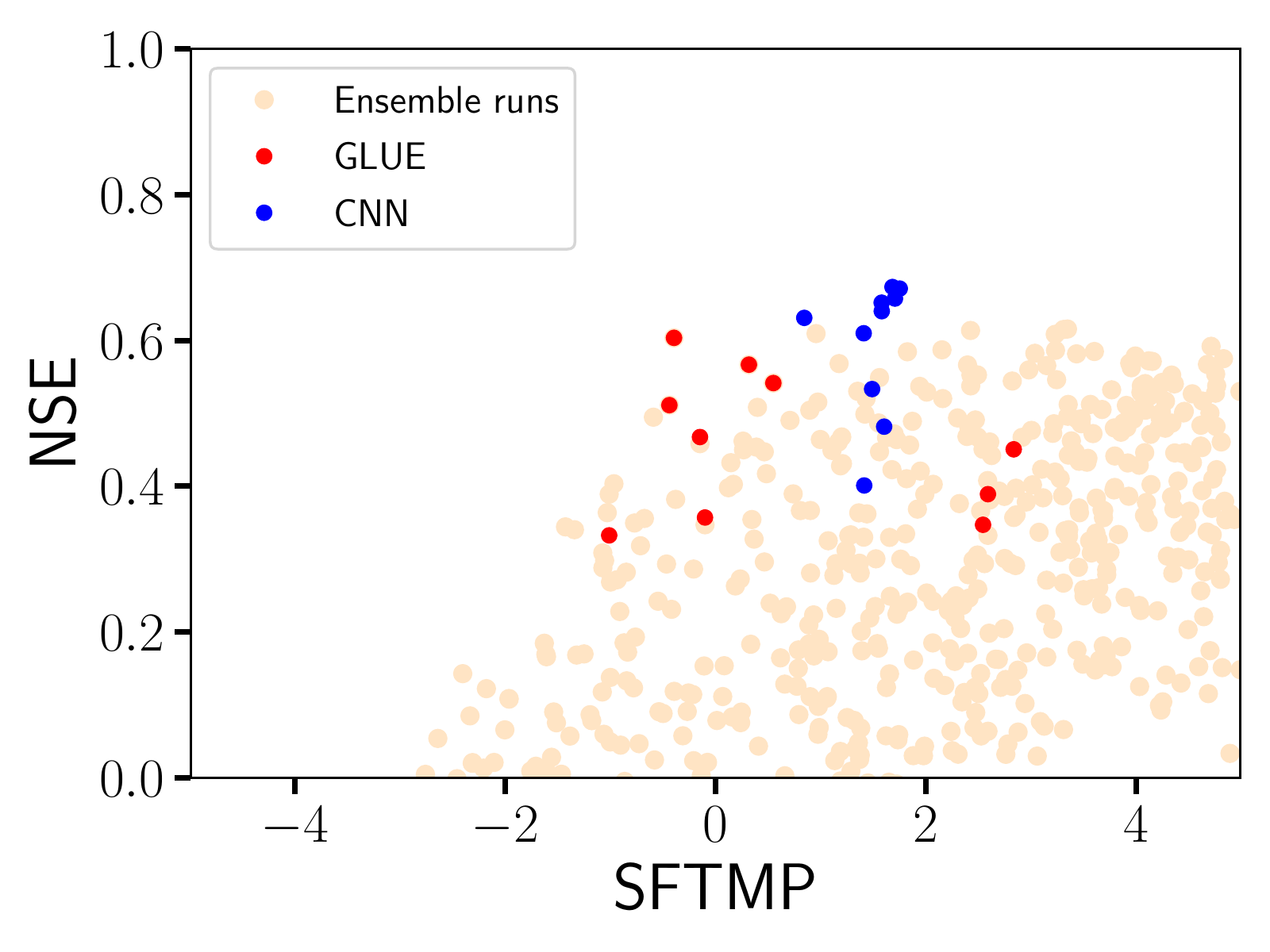}}
    \subfigure[logNSE vs. SFTMP]
    {\includegraphics[width = 0.475\textwidth]{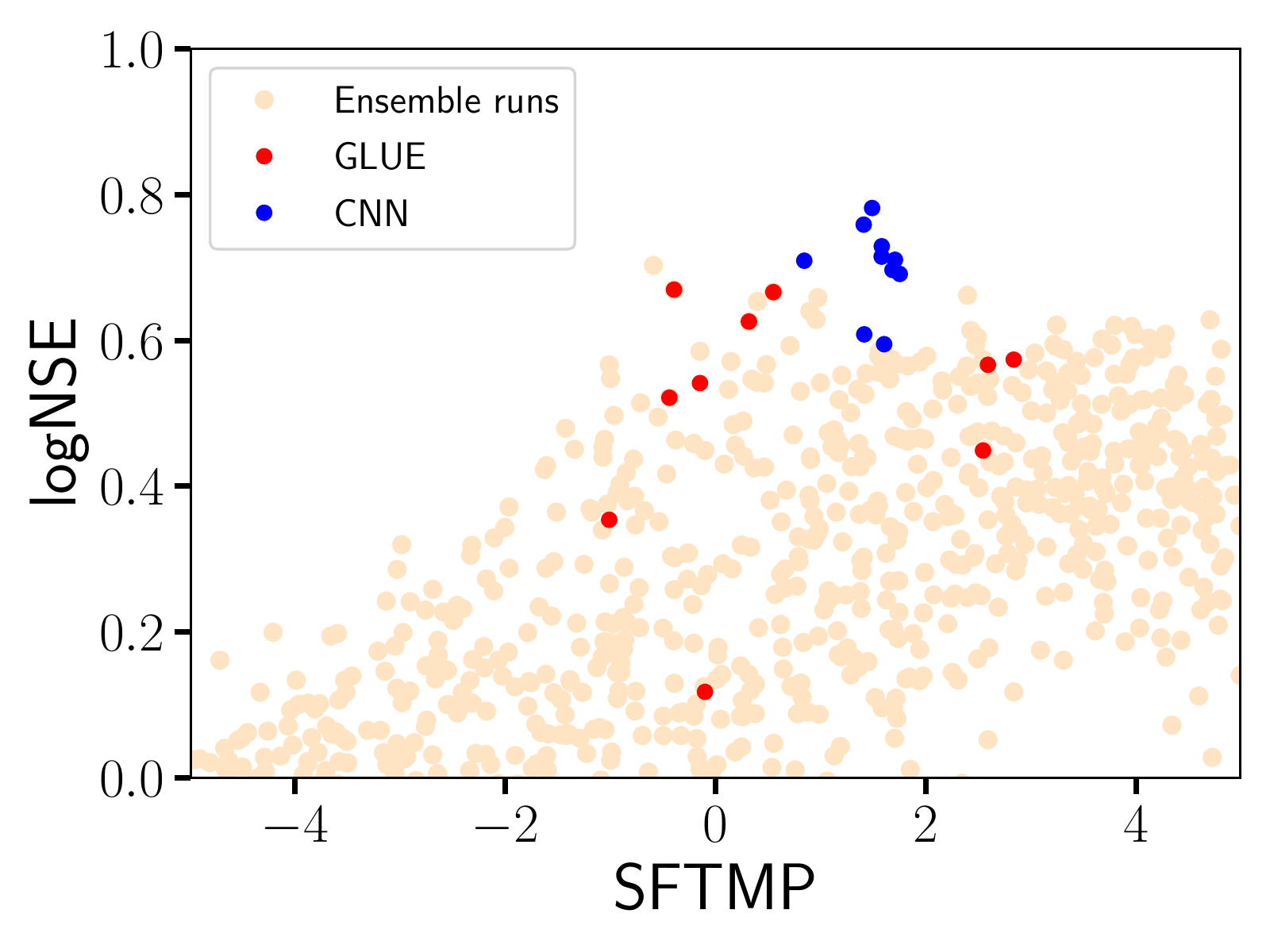}}
  \caption{\textbf{Estimated SWAT parameter (SFTMP) using observational data (DL vs. GLUE):}~This figure compares the estimation of the SFTMP parameter using observational data with four different performance metrics.
  The calibration sets are identified using the top 10 DL-based inverse models and the GLUE method.
  The blue, red, and light orange colored data points represent the DL estimation, GLUE behavioral set, and ensemble runs in parameter space.
  Across performance metrics, it is evident that estimation using the DL-enabled inverse models is better clustered than GLUE.
  \label{Fig:DL_vs_GLUE_Obs_Data_Top7Params}}
\end{figure}

Figure~\ref{Fig:DL_vs_GLUE_Obs_Data_Top7Params} shows the DL estimation of SFTMP and demonstrates its calibration performance with four different metrics.
Supplementary Figs.~S17-S22 provide estimations for the other six sensitive parameters.
We compare the DL-enabled estimations with estimates from the GLUE-based method and prior ensemble predictions.
The evaluation criteria include $R^2$-score, NSE, logNSE, and KGE.
For instance, NSE, logNSE, and KGE are evaluated as follows:
\begin{subequations}
  \begin{align}
    \label{eq:NSE}
    \mathrm{NSE}(\mathbf{q},\hat{\mathbf{q}}) &= 1 - \frac{\displaystyle \sum \limits_{i = 1}^{n} 
    \left(q_i -\hat{q_i} \right)^2}{\displaystyle \sum \limits_{i = 1}^{n} 
    \left(q_i - \mu_{\mathbf{q}} \right)^2} \quad \mathrm{where} \; \; \mu_{\mathbf{q}} = \frac{1}{n} \displaystyle 
    \sum \limits_{i = 1}^{n} q_i \\
    \mathrm{logNSE}(\mathbf{q},\hat{\mathbf{q}}) &= 1 - \frac{\displaystyle \sum \limits_{i = 1}^{n} 
    \left(\log{[q_i]} - \log{[\hat{q_i}]} \right)^2}{\displaystyle \sum \limits_{i = 1}^{n} 
    \left(\log{[q_i]} - \log{[\bar{q}]} \right)^2} \\
    \mathrm{KGE}(\mathbf{q},\hat{\mathbf{q}}) &= 1 - 
    \sqrt{\left(r - 1\right)^2 + 
    \left(\frac{\sigma_{\hat{\mathbf{q}}}}{\sigma_{\mathbf{q}}} - 1\right)^2 + 
    \left(\frac{\mu_{\hat{\mathbf{q}}}}{\mu_{\mathbf{q}}} - 1\right)^2}
  \end{align}
\end{subequations}
where $\hat{q_i} \in \hat{\mathbf{q}}$ is the \texttt{SWAT} model predictions and $q_i \in \mathbf{q}$ is the observational streamflow.
$n$ is the dimension of $\hat{\mathbf{q}}$ and $\mathbf{q}$, which is the total number of time-steps.
$r$ is the Pearson product-moment correlation coefficient.
$\sigma_{\hat{\mathbf{q}}}$ and $\sigma_{\mathbf{q}}$ are the standard deviations in the \texttt{SWAT} model predictions and observations, respectively.
$\mu_{\hat{\mathbf{q}}}$ and $\mu_{\mathbf{q}}$ are the standard deviations in the \texttt{SWAT} model predictions and observations, respectively.

Each metric takes into account different aspects of calibration performance \cite{liu2020rational}.
The $R^2$-score indicates the goodness of fit, which measures how close the streamflow predictions from the DL-enabled calibration are to the observed data.
NSE evaluates how well the calibrated \texttt{SWAT} model predictions capture high flows.
Complementary to NSE, logNSE determines the accuracy of model predictions concerning low flows.
KGE combines these three different components of NSE (i.e., correlation, bias, a ratio of variances or coefficients of variation) in a more balanced way (e.g., more weight on low flows and less weight on extreme flows) to assess the \texttt{SWAT} model calibration.
To summarize, DL-enabled parameter estimation is better than the behavioral parameter sets estimated by GLUE for all four studied metrics.
Moreover, the uncertainty ranges in DL estimated parameters are low compared to GLUE and result in a well-clustered set.

\begin{figure}
  \centering
    \subfigure[Calibration period]{\includegraphics[width = 0.485\textwidth]{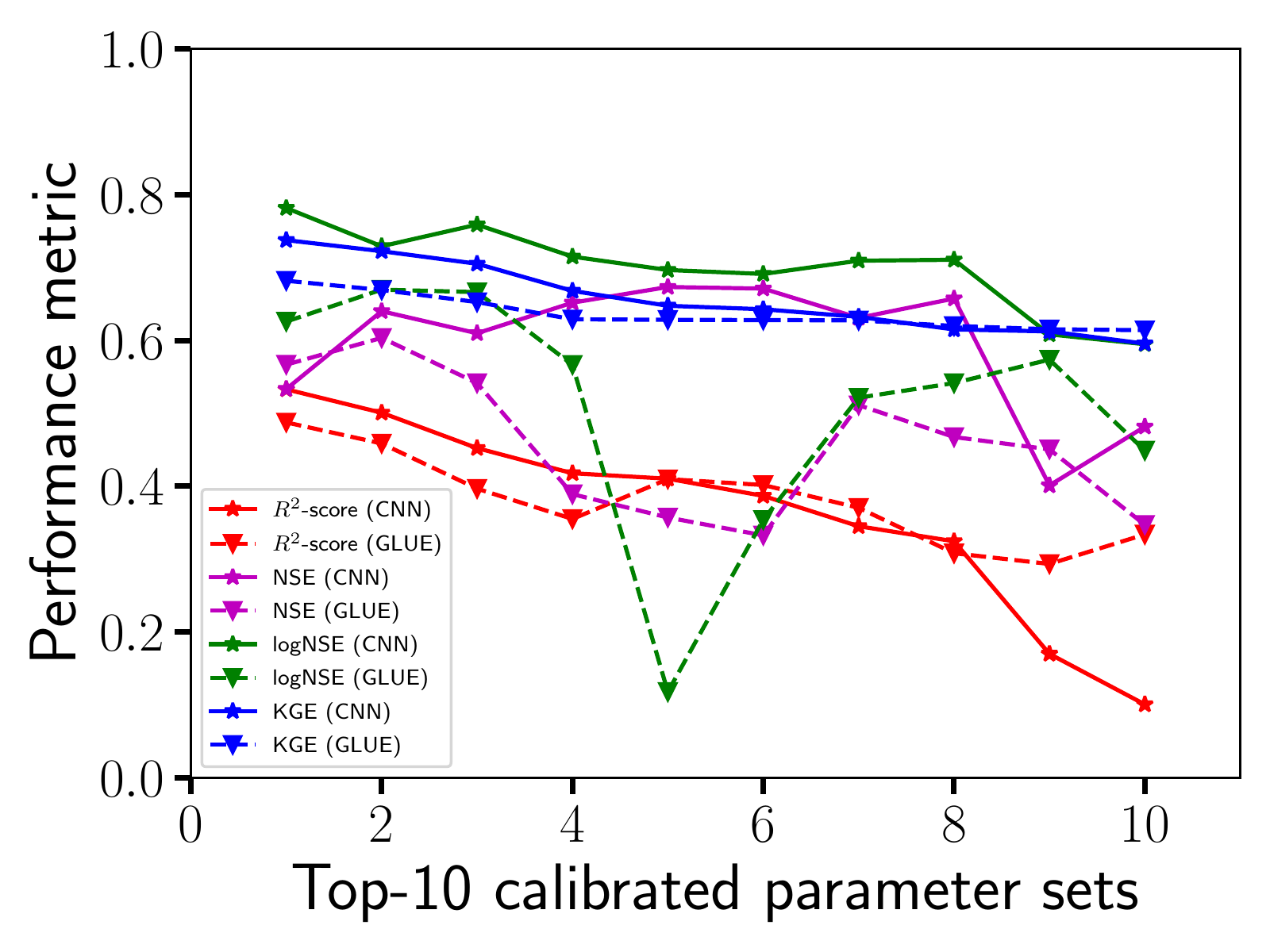}}
    \subfigure[Validation period]{\includegraphics[width = 0.485\textwidth]{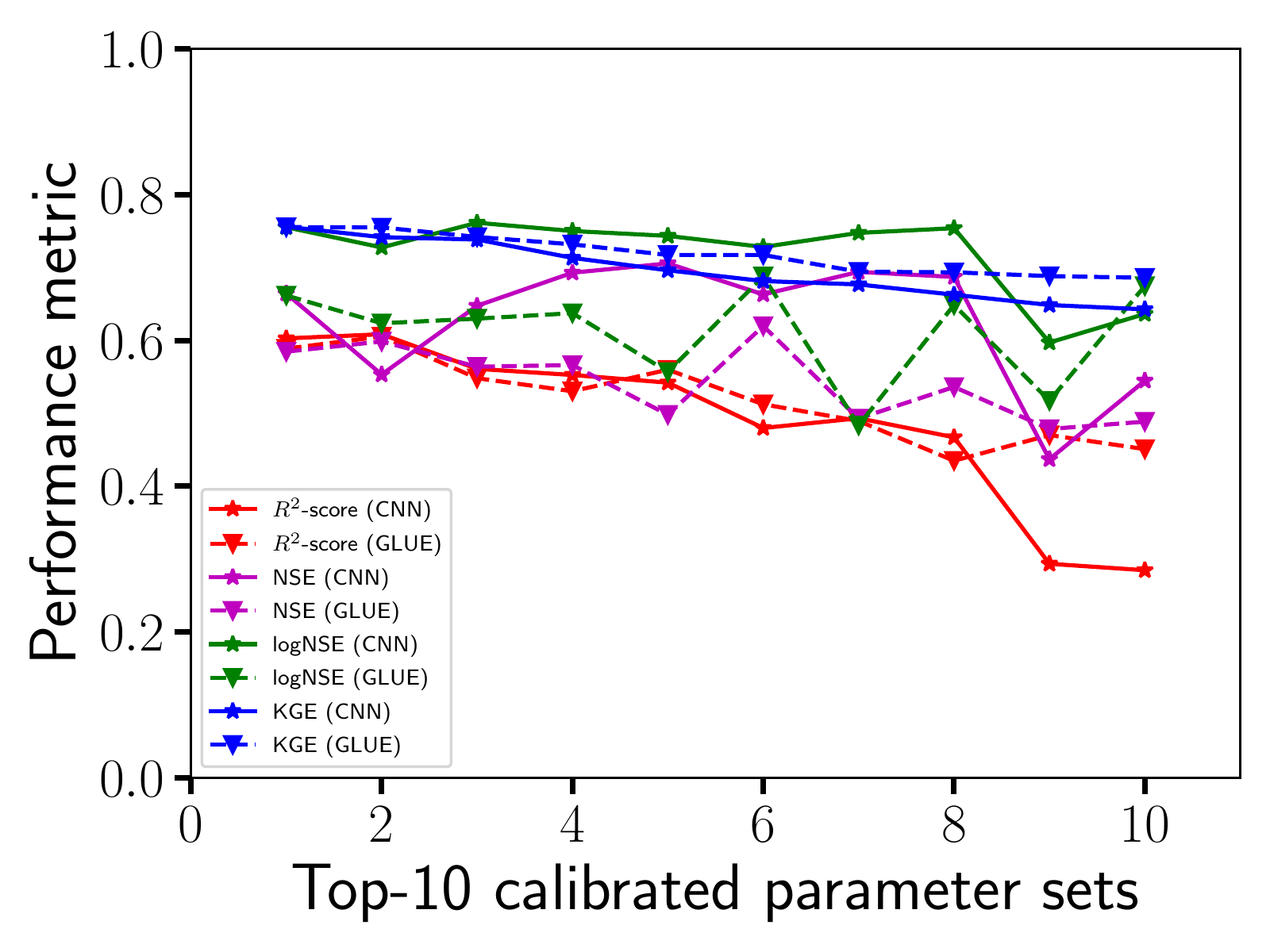}}
    \subfigure[{\scriptsize Calibration period:~KGE-based best set (KGE = 0.74)}]
    {\includegraphics[width = 0.245\textwidth]{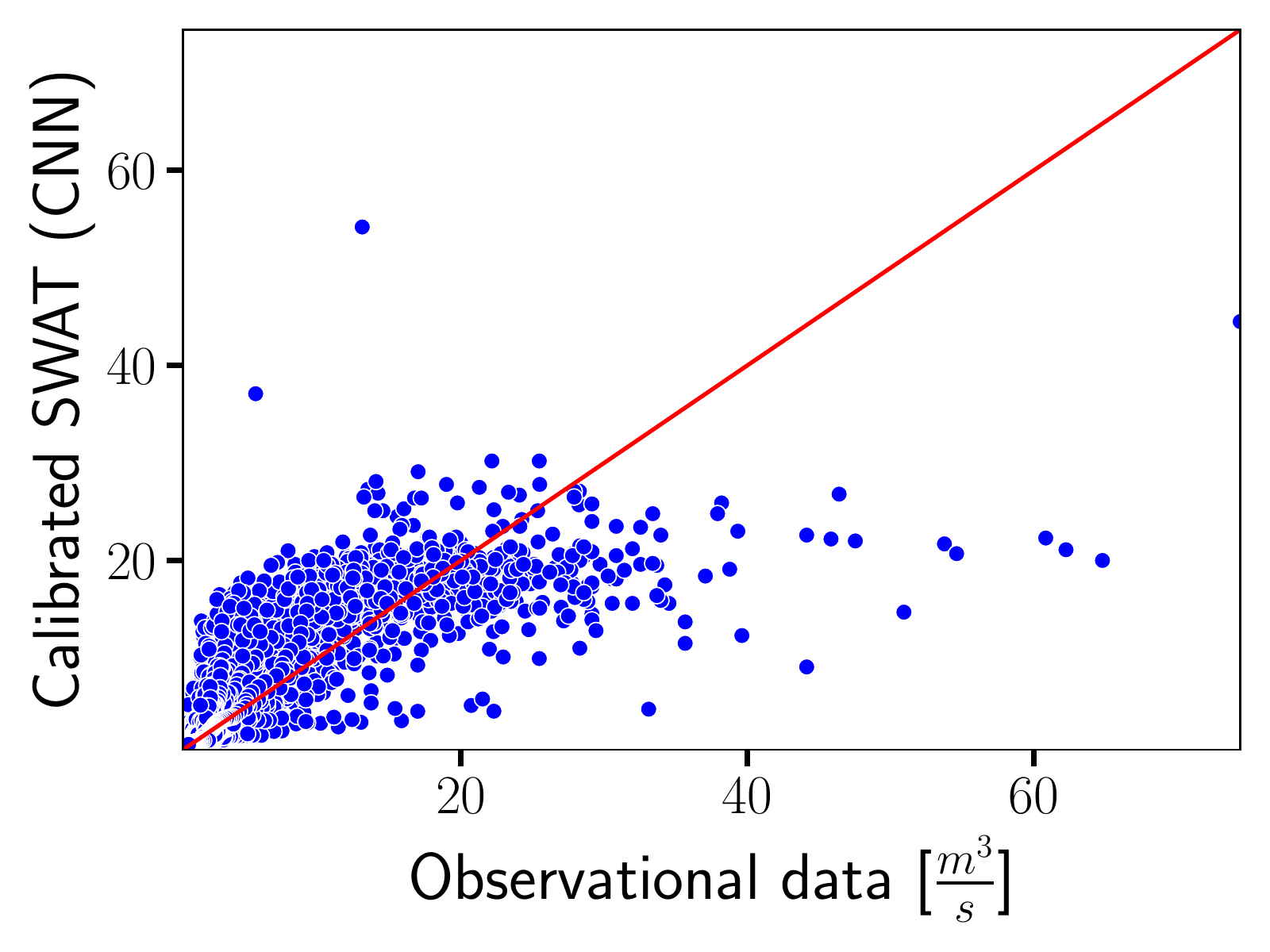}}
    \subfigure[{\scriptsize Validation period:~KGE-based best set (KGE = 0.76)}]
    {\includegraphics[width = 0.245\textwidth]{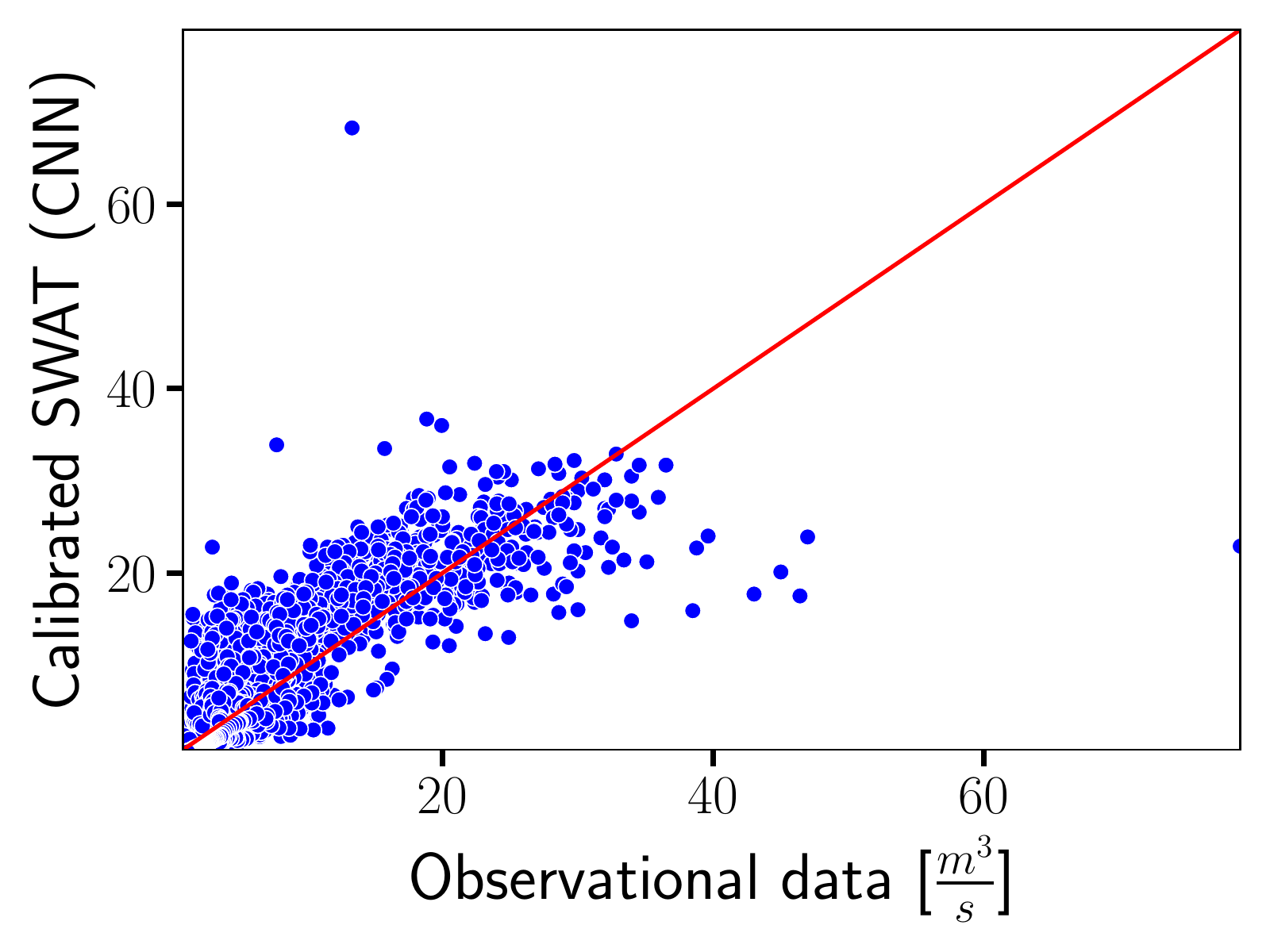}}
    \subfigure[{\scriptsize Calibration period:~KGE-based best set (KGE = 0.68)}]
    {\includegraphics[width = 0.245\textwidth]{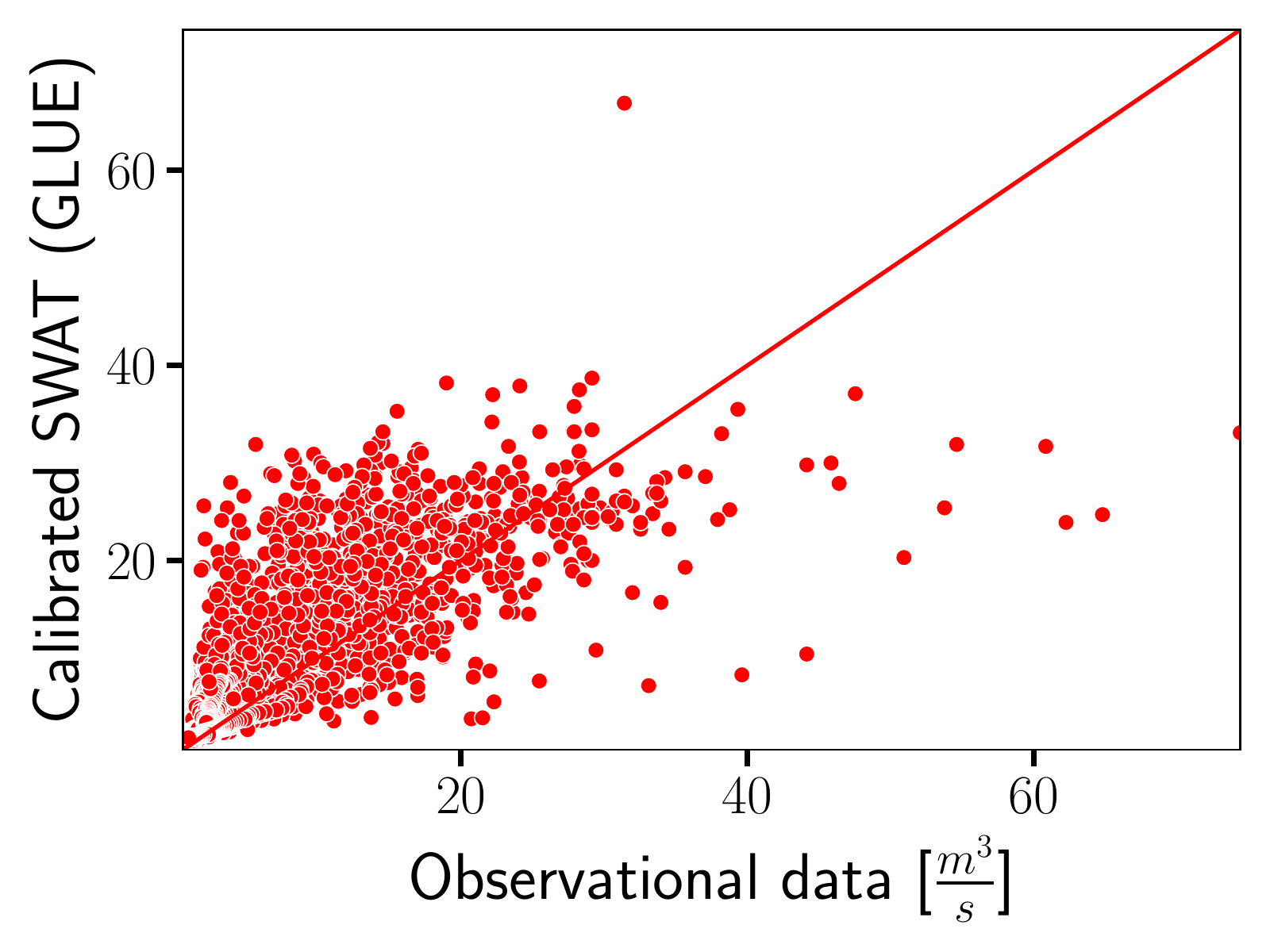}}
    \subfigure[{\scriptsize Validation period:~KGE-based best set (KGE = 0.75)}]
    {\includegraphics[width = 0.245\textwidth]{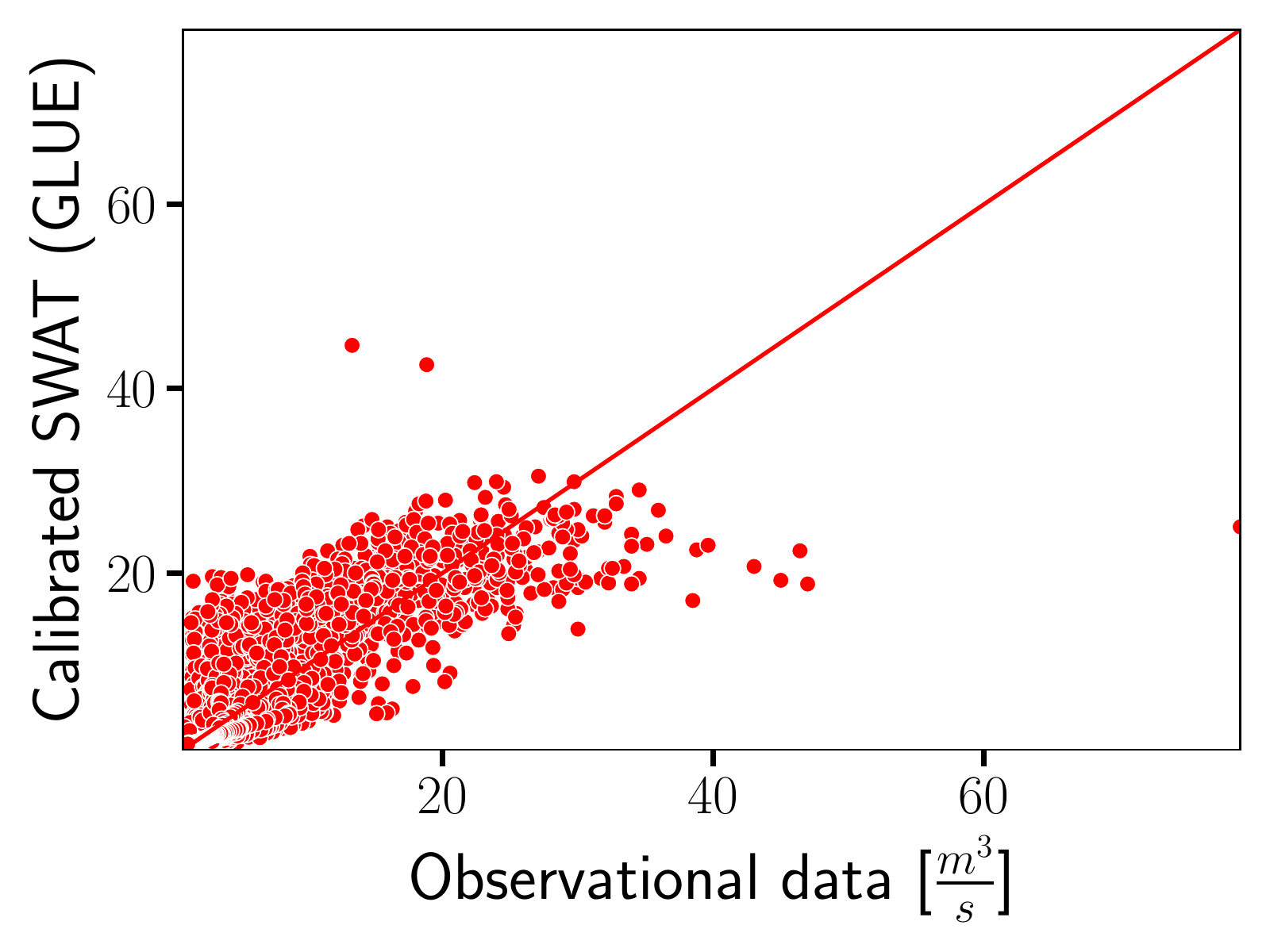}}
    \subfigure[{\scriptsize Size of mean variation of streamflow vs. time period}]
    {\includegraphics[width = 0.385\textwidth]{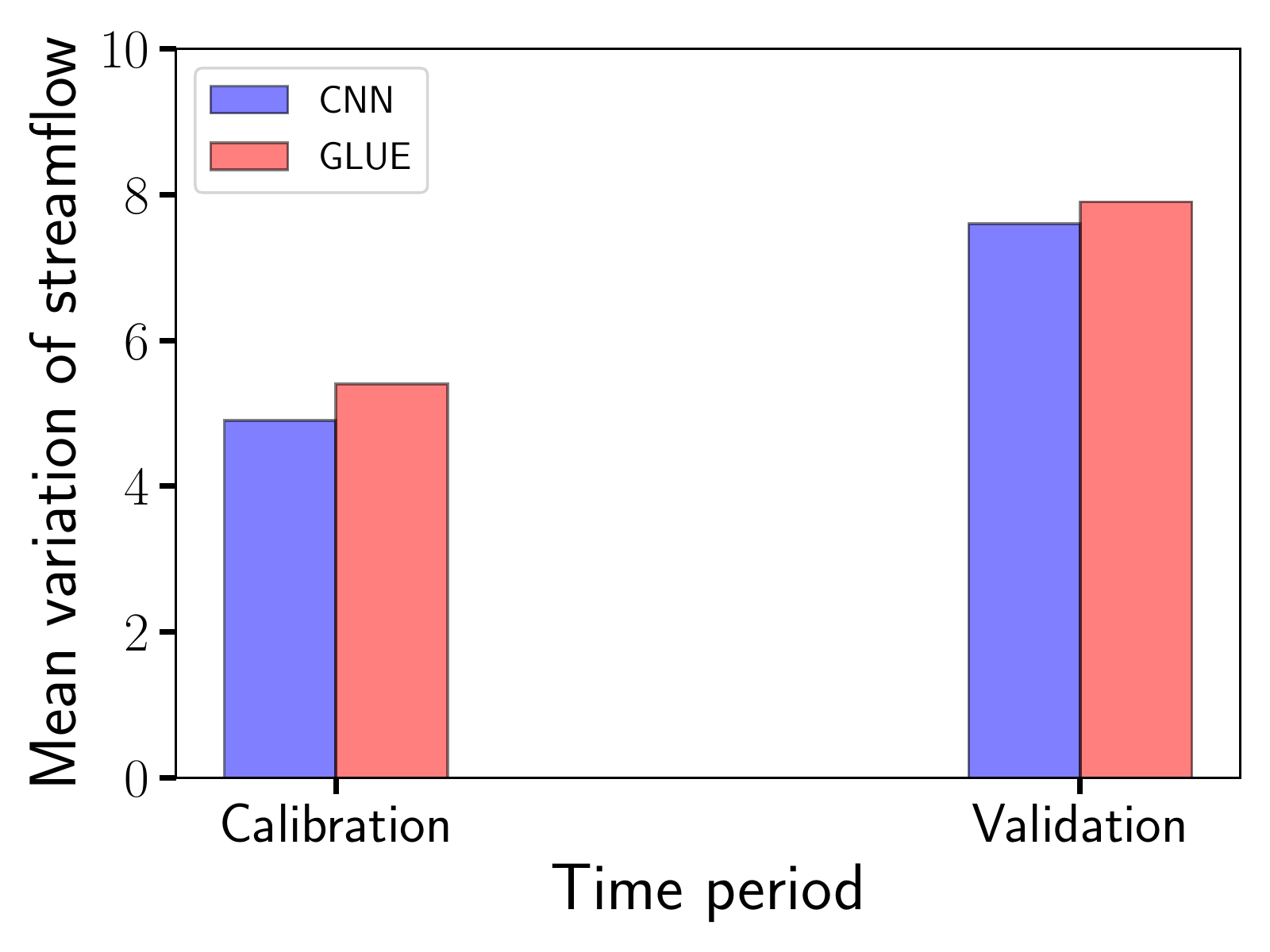}}
    \subfigure[{\scriptsize Probability that observational data is contained within the prediction bounds estimated by DL or GLUE}]
    {\includegraphics[width = 0.385\textwidth]{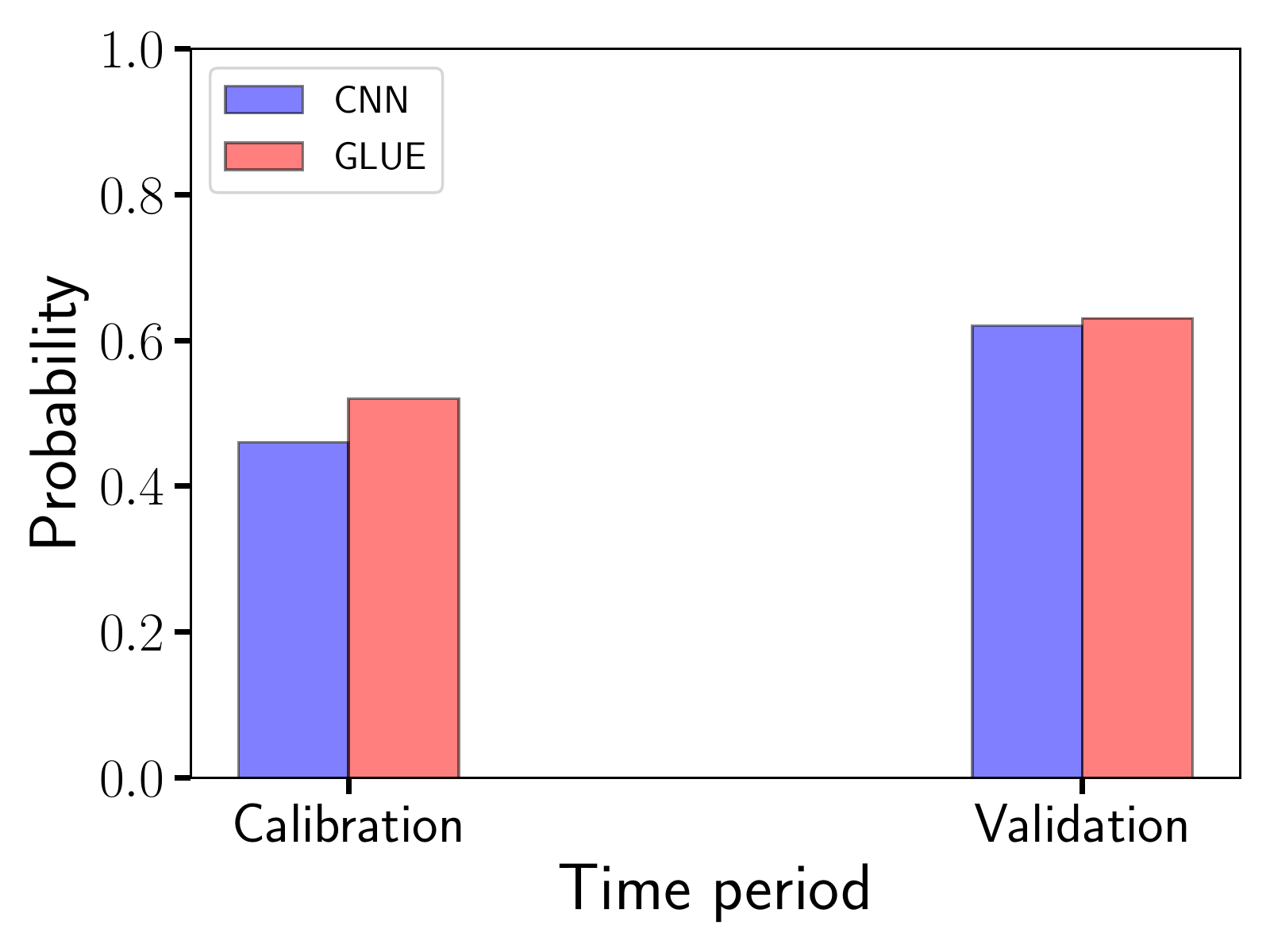}}
  \caption{\textbf{Performance of DL vs. GLUE estimated parameters on observational data:}~This figure compares the performance of the calibrated parameter sets evaluated by DL (i.e., using the 21-parameter inverse model) and GLUE within and beyond the calibration period.
  The top figures show the performance metrics of the top 10 sets in the calibration and validation period.
  The performance metrics (e.g., NSE, logNSE) focus on the predictive capability of DL- and GLUE-based calibrated \texttt{SWAT} models in both low and high flow scenarios.
  DL outperforms GLUE across all metrics.
  The central figures provide the one-to-one plots of the calibrated \texttt{SWAT} model compared to observational discharge.
  The discharge predictions are based on the best calibrated set from the DL-enabled inverse model and GLUE methods.
  The red line is the one-to-one line.
  Data points based on the DL estimated set are closer to the red line than data from the GLUE-based set. 
  The bottom figure shows the size of the mean streamflow variation.
  It also provides the probability that the observational data is contained within the prediction bounds (i.e., the light blue or red colored region in Fig.~\ref{Fig:DL_vs_GLUE_CalibTimePeriod}) estimated by the calibrated \texttt{SWAT} model.
  The uncertainty in the GLUE-based calibration sets prediction, and associated probability is higher than DL.
  \label{Fig:DL_vs_GLUE_Performance_Metrics}}
\end{figure}

Figure~\ref{Fig:DL_vs_GLUE_Performance_Metrics}(a) and (b) compares the predictive performance of the top ten calibrated parameter sets within and beyond the calibration period.
These figures show that most of the DL estimations consistently perform better than the GLUE-based sets.
The one-to-one scatter plots in Figs.~\ref{Fig:DL_vs_GLUE_Performance_Metrics}(c)-(f) compare the streamflow predictions with observational data, where each dot corresponds to daily streamflow.
The predictions are based on the best sets calibrated by either DL or GLUE.
The best DL-based calibrated set has $R^2$, NSE, logNSE, and KGE scores of 0.53, 0.67, 0.87, and 0.74, respectively.
The best GLUE-based calibrated set has $R^2$, NSE, logNSE, KGE scores of 0.48, 0.6, 0.7, and 0.68, respectively.
From these values, it is clear that the DL-enabled inverse model estimations have higher accuracy in the \texttt{SWAT} model calibration than the GLUE estimations.
Therefore, CNNs show promise for parameter estimation, especially in nonlinearly relating streamflow data to conceptual parameters.

\begin{figure}
  \centering
    \subfigure[Calibration period:~Calibrated \texttt{SWAT} model predictions based on DL estimated parameters]
    {\includegraphics[width = 0.985\textwidth]{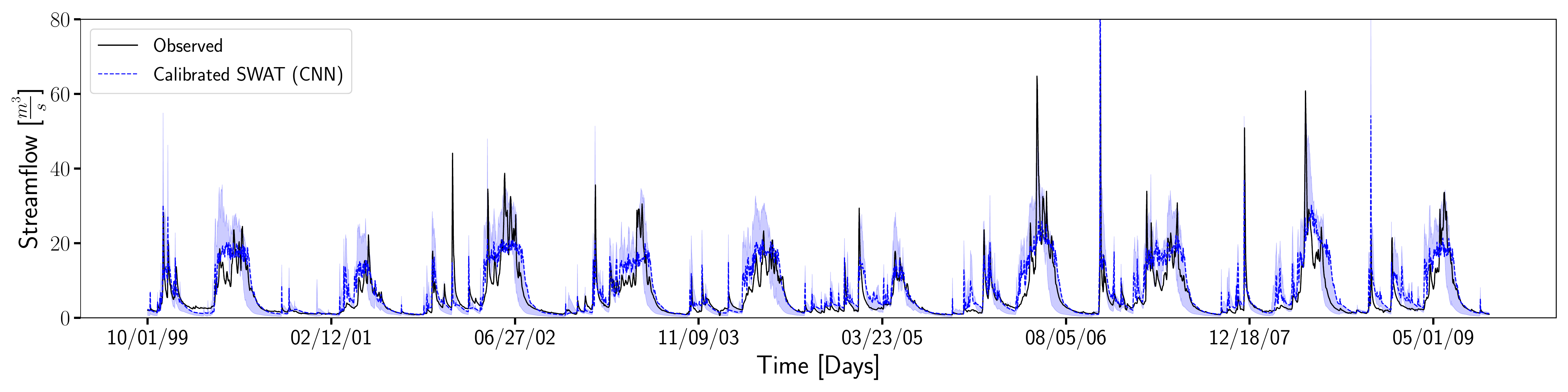}}
    \subfigure[Calibration period:~Calibrated \texttt{SWAT} model predictions based on GLUE estimated parameters]
    {\includegraphics[width = 0.985\textwidth]{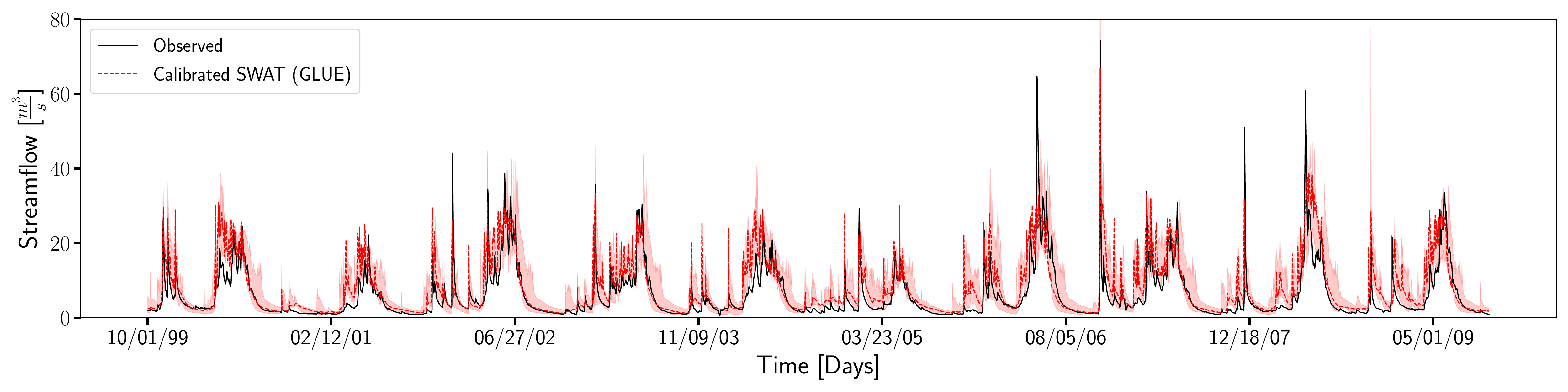}}
    \subfigure[Validation period:~Calibrated \texttt{SWAT} model predictions based on DL estimated parameters]
    {\includegraphics[width = 0.985\textwidth]{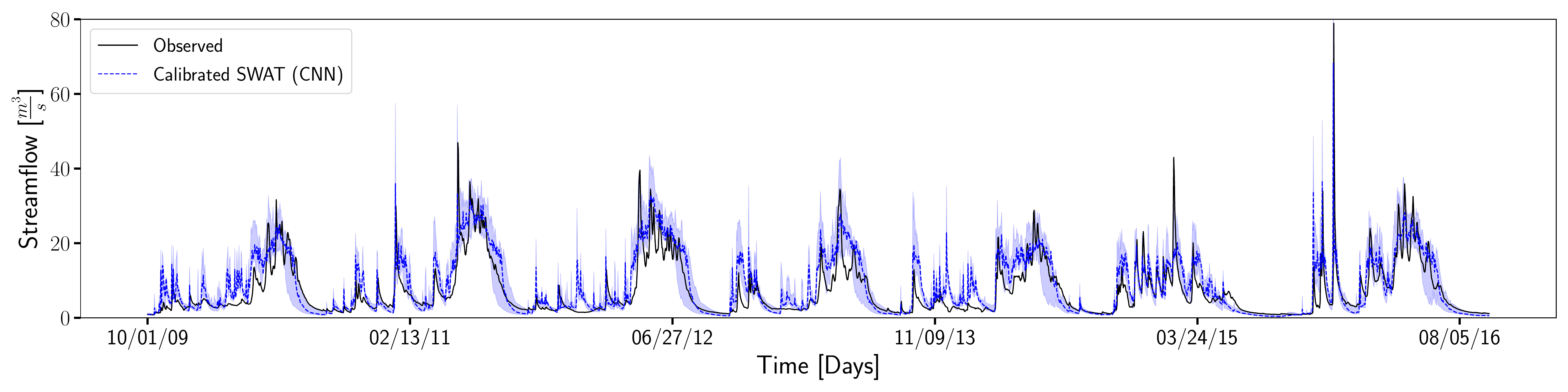}}
    \subfigure[Validation period:~Calibrated \texttt{SWAT} model predictions based on GLUE estimated parameters]
    {\includegraphics[width = 0.985\textwidth]{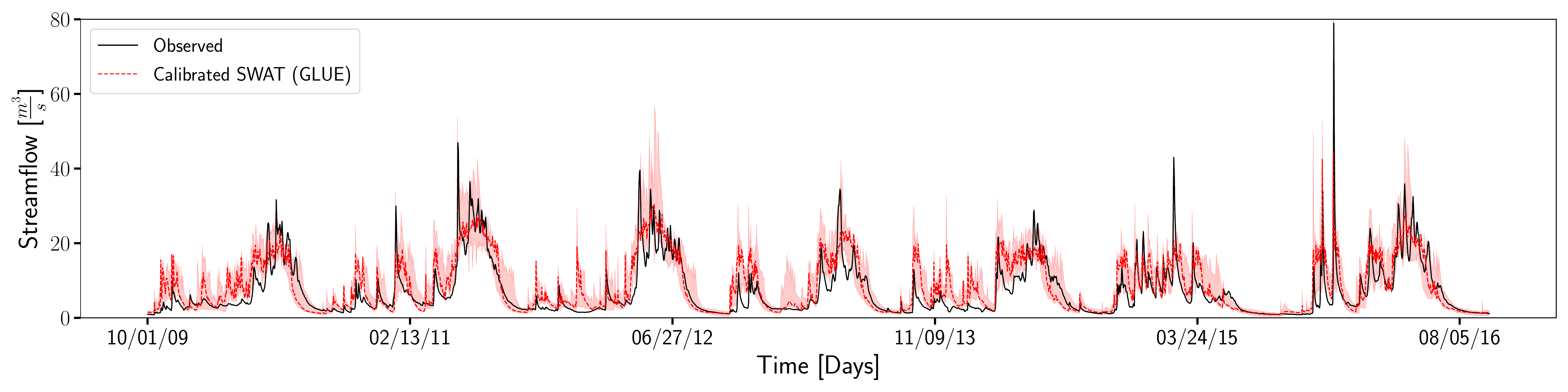}}
  \caption{\textbf{Comparison of the calibrated \texttt{SWAT} model (DL vs. GLUE) with observation data:}~This figure compares the predictions of the calibrated \texttt{SWAT} model with observational data within and beyond the calibration period.
  The solid black line represents the observational data.
  The dashed colored (blue or red) line represents the predictions based on the best calibrated set using DL (i.e., estimation based on 21-parameter inverse model) or GLUE.
  The light colored region (blue or red) represents the prediction uncertainty. This region is calculated by running the \texttt{SWAT} model using the remaining nine calibration sets obtained using DL or GLUE.
  \label{Fig:DL_vs_GLUE_CalibTimePeriod}}
\end{figure}

Figure~\ref{Fig:DL_vs_GLUE_Performance_Metrics}(g) shows that the uncertainty ranges are narrower for the top ten DL estimations in both calibration and validation periods than the GLUE estimations.
As a result, the probability that the prediction intervals estimated by the DL sets contain the observed streamflow is lower compared to GLUE, as seen in Fig.~\ref{Fig:DL_vs_GLUE_Performance_Metrics}(h).
One of our next steps is to improve this predictive uncertainty through probabilistic Bayesian neural networks \cite{lu2021streamflow}.
This type of network can account for uncertainty so that DL-enabled inverse models can assign lesser confidence levels to incorrect predictions.
Figure.~\ref{Fig:DL_vs_GLUE_CalibTimePeriod} compares the streamflow predictions from the calibrated \texttt{SWAT} with the observed data using the top ten DL and GLUE estimated sets.
The DL estimations capture the various high and low flows better than GLUE in both the calibration and validation periods.
However, the calibrated \texttt{SWAT} model demonstrates over predictions in the year 2008 and under forecasts in the year 2005.
This lower predictive performance may imply potential deficiencies (i.e., structural errors) in the underlying \texttt{SWAT} model representation of the watershed processes. 
Additional investigations are necessary to identify other processes and parameters that reduce structural errors and discrepancies in streamflow predictions.

\subsection{Discussion}
\label{SubSec:S4_Discussion}
Our results demonstrate the applicability of using deep learning to calibrate the \texttt{SWAT} model.
We note that the proposed methodology is general and can be used to calibrate other watershed models such as \texttt{ATS} and \texttt{PRMS}.
This extensibility for calibrating other models and study sites can be achieved using transfer learning methods \cite{zhuang2020comprehensive}.
Transfer learning will allow us to reuse the CNNs developed in this study and leverage them for a new, similar problem.
Minimal re-training is necessary to fine-tune the trained CNNs and apply them to calibrate watershed models for other study sites.
Such a transfer of knowledge across study sites is usually performed when generating large training data needed to develop a full-scale CNN and tune its trainable weights from the start is too computationally expensive (e.g., when using \texttt{ATS}).
Additionally, we can improve our DL methodology to calibrate the \texttt{SWAT} model by incorporating other multi-source data streams (e.g., ET, SWE) along with streamflow.
Our next step is to use such data streams to further investigate the deficiency of the model structure or processes in the \texttt{SWAT} by ingesting streamflow, ET, and SWE into CNNs.

Figures~\ref{Fig:DL_vs_GLUE_Performance_Metrics}(g) and (h) show that the uncertainty boundary of the streamflow predictions estimated using the DL sets are narrower than GLUE, which is a limitation of the current methodology.
Our future work involves investigating this DL-based predictive uncertainty.
Improved uncertainty intervals can be calculated by performing ensemble \texttt{SWAT} model runs using the calibration sets (i.e., 500 runs for a single tuned CNN) with all the observational errors (i.e., from 5\% to 25\%).
Specifically, computing streamflow predictive uncertainty with noise involves running a total of 5000 DL estimated calibration sets, which may be computationally expensive.
However, incorporating the DL calibration sets obtained using noisy observational data may allow us to better understand the uncertainty intervals of streamflow predictions \cite{kiang2018comparison}. 
In addition to observational errors, developing CNNs tailored to estimate the \texttt{SWAT} model parameters under different hydrological seasons \cite{mcmillan2020linking} (e.g., winter vs. summer) may enhance the calibration process.
For example, comparing DL estimated sets from wet and dry periods of the year can provide better insights on the \texttt{SWAT} model parameters that control streamflow predictions across different seasons.
When making such comparisons between real data and model predictions, hydrological signatures and their associated metrics \cite{westerberg2015uncertainty,mcmillan2017five,mcmillan2021review,fatehifar2021assessing,gnann2021tossh} can be used to shed light on the structural deficiencies of the \texttt{SWAT} model. 
Hydrological signatures on which we can evaluate performance metrics include the slope of flow duration curve, rising limb density, recession shape, and baseflow index of streamflow time-series data \cite{mcmillan2021review}.

To conclude our discussion, in addition to the data-driven methodology presented in this study\footnote{or by combining MCMC with forward emulators \cite{dagon2020machine} for model calibration}, the efficacy of the proposed DL methodology can also be improved by embedding domain knowledge into deep neural networks \cite{read2019process,khandelwal2020physics,bhasme2021enhancing,jia2021physics}.
Recent advances in knowledge-guided machine learning provide an avenue to incorporate model states/fluxes and water balance as part of recurrent neural network architectures \cite{khandelwal2020physics}.
The papers mentioned above used such neural architectures to develop forward emulators for watershed models.
One can extend the methods presented in those works to incorporate process model knowledge into our proposed CNNs to improve \texttt{SWAT} model calibration.

\section{CONCLUSIONS}
\label{Sec:S5_Conclusions}
In this paper, we have developed a fast, accurate, and reliable methodology better to calibrate the \texttt{SWAT} model.
The developed DL-enabled inverse models were used to estimate \texttt{SWAT} parameters for the ARW study site in YRB.
Our approach leverages recent advances in deep learning techniques, CNNs, to extract representations from streamflow data and map them to \texttt{SWAT} model parameters.
Hyperparameter tuning was performed to identify optimal CNN architectures.
Ensemble runs from the \texttt{SWAT} model are used to train, validate, and test three different DL-enabled inverse models (i.e., 1-parameter, 7-parameter, and 21-parameter models).
We also used sensitivity analysis of the ensemble runs to identify the dominant parameters that influence streamflow.
Two of three DL-enabled inverse models (i.e., 1-parameter and 7-parameter models) were developed to identify the crucial parameters.
Our results show that all three DL models were able to estimate the sensitive parameters reasonably well.
Specifically, the 1-parameter model performed slightly better than the 7-parameter and 21-parameter models.
The parameters estimated from all three models were robust to high observational errors.
We then compared the \texttt{SWAT} parameters estimated by DL with those generated by the traditional GLUE method.
We found that the DL estimated \texttt{SWAT} parameters are within the sampling range of the ensemble runs and are well clustered compared to the GLUE. 
Furthermore, this comparison also showed that the predictions of the calibrated \texttt{SWAT} model based on CNNs consistently outperformed the GLUE method.
Key performance metrics (e.g., NSE and logNSE) showed that the DL-based calibration sets capture low and high flows better than GLUE. 
This improvement in predictive performance is because the CNNs can more effectively utilize the information (e.g., learning representative features from streamflow) provided in ensemble runs than the GLUE method. 
By capturing the nonlinear relationships between the \texttt{SWAT} model inputs and outputs through multiple convolutional neural layers, the CNNs yielded more realistic parameter estimations for the ARW and a better calibrated \texttt{SWAT} model.
This improvement resulted in a closer match between model-predicted and observed stream discharges.

From a computational cost perspective, the time needed 
to infer parameters based on DL is at least $\mathcal{O}(10^3)$ faster than GLUE, which makes extending this workflow to complex watershed models (e.g., \texttt{ATS}) attractive.
The training time needed to develop DL models can further be improved by using GPUs and tensor processing units \cite{bisong2019google}.
Reducing the computational cost of developing DL-enabled inverse models is one of our next steps, with a focus on using the distributed training (e.g., using \textsf{Horovod} \cite{sergeev2018horovod} or \textsf{DeepHyper} \cite{balaprakash2018deephyper}) that already shows promise for achieving this type of speedup.
This improves the efficiency during training by splitting the workload of training the CNNs among multiple processors.
Our methodology is general and can be utilized to calibrate complex watershed models (i.e., through transfer learning methods \cite{zhuang2020comprehensive}) with minimal re-training.
Additional future work is modifying the proposed workflow to incorporate multi-source datasets (e.g., by combining streamflow, ET, SWE) to further enhance \texttt{SWAT} model calibration.

\section*{ABBREVIATIONS}
\begin{itemize}
  \item ARW: American River Watershed 
  \item \texttt{ATS}:~Advanced Terrestrial Simulator
  \item CN:~Curve number
  \item CNN: Convolutional Neural Network 
  \item \texttt{DART}:~The Data Assimilation Research Testbed
  \item \texttt{DHSVM}:~The Distributed Hydrology Soil Vegetation Model
  \item DEM:~Digital Elevation Model
  \item DNN:~Deep Neural Network
  \item DL: Deep Learning 
  \item ET:~Evapotranspiration
  \item FAIR:~Findable Accessible Interoperable and Reusable
  \item GIS:~Geographic information System
  \item GLUE:~Generalized Likelihood Uncertainty Estimation
  \item GSA:~Global Sensitivity Analysis
  \item GPU:~Graphical Processing Unit
  \item HRU:~Hydrologic Response Unit
  \item \texttt{HSPF}:~Hydrological Simulation Program-Fortran
  \item \texttt{MADS}:~Model Analysis \& Decision Support
  \item KGE:~Kling-Gupta Efficiency
  \item \texttt{MATK}:~Model Analysis ToolKit
  \item ModEx:~Modeling-Experimental approach
  \item MTL:~Multi-task Learning
  \item NLCD:~National Land cover Database
  \item NHDPlus:~National Hydrography Dataset Plus
  \item NSE:~Nash-Sutcliffe efficiency
  \item logNSE:~logarithmic Nash-Sutcliffe Efficiency
  \item \texttt{NWM}:~The National Water Model
  \item PET:~Potential Evapotranspiration
  \item \texttt{PEST}:~Parameter Estimation software
  \item PRISM:~Parameter Elevation Regression on Independent Slopes Model
  \item \texttt{PRMS}:~Precipitation Runoff Modeling System
  \item \texttt{RHESSys}:~Regional Hydro-Ecologic Simulation System
  \item SCE-UA:~Shuffled Complex Evolution Method developed at The University of Arizona
  \item \texttt{SWAT}:~Soil and Water Assessment Tool
  \item \texttt{SWAT-CUP}:~\texttt{SWAT} Calibration and Uncertainty Programs
  \item SNOTEL:~Snow Telemetry
  \item STATSGO:~Soil Maps for the State Soil Geographic
  \item STL:~Single-task Learning
  \item TPU:~Tensor Processing Unit
  \item USGS:~United States Geological Survey
  \item \texttt{WRF-Hydro}:~The Weather Research and Forecasting Model Hydrological Modeling System
  \item \texttt{VIC}:~The Variable Infiltration Capacity model
  \item YRB:~Yakima River Basin
\end{itemize}

\section*{ACKNOWLEDGMENTS}
This research was supported by the U.S. Department of Energy (DOE), Office of Science (SC) Biological and Environmental Research (BER) program, as part of BER's Environmental System Science (ESS) program. 
This contribution originates from the River Corridor Scientific Focus Area (SFA) at Pacific Northwest National Laboratory (PNNL).
This research used resources from the National Energy Research Scientific Computing Center (NERSC), a DOE-SC User Facility. 
The views and opinions of authors expressed herein do not necessarily state or reflect those of the United States Government or any agency thereof.

\section*{\textbf{CONFLICT OF INTEREST}}
The authors declare that they do not have any conflicts of interest.

\section*{\textbf{APPENDIX (supplementary materials)}}
Supplementary materials and associated figures can be found in a separate file.

\section*{\textbf{DATA, CODE AVAILABILITY, AND RESOURCES}}
The data generated for the proposed deep learning model development uses open-science principles.
Specifically, we make the data Findable Accessible Interoperable and Reusable (FAIR).
FAIR principles expedite community-based data generation, modeling, interdisciplinary collaboration and provides a means to test new hypotheses.
The datasets generated and analyzed as well as scripts for this study, will be made available on this PNNL Gitlab repository:~\url{https://gitlab.pnnl.gov/sbrsfa/dnn_swat_calib.git}.
\texttt{SWAT} open-source code can be downloaded at \url{https://swat.tamu.edu/}.

\bibliographystyle{unsrt}
\bibliography{Master_References/SWAT_Calib_References}

%
\end{document}